\pdfoutput=1
\documentclass[11pt]{article}
\PassOptionsToPackage{table}{xcolor}   %
\usepackage[preprint]{acl}
\makeatletter   %
\let\WSP@thebibliography\thebibliography
\renewcommand{\thebibliography}[1]{\WSP@thebibliography{#1}\raggedright}
\makeatother
\usepackage{times}
\usepackage{latexsym}
\usepackage[T1]{fontenc}
\usepackage[utf8]{inputenc}
\usepackage{microtype}
\usepackage{inconsolata}
\usepackage{graphicx}
\usepackage{booktabs}
\usepackage{amsmath,amssymb}
\usepackage{xcolor}
\usepackage{tabularx}
\usepackage{enumitem}
\usepackage{tikz}
\usetikzlibrary{positioning}
\usepackage{array}
\setlist{nosep,leftmargin=1.2em}
\newcommand{\nCollFt}{4}
\newcommand{\nCollFtN}{264}
\newcommand{\nCollLora}{20}
\newcommand{\nCollLoraN}{264}
\newcommand{\nCollPft}{0}

\newcommand{\nCollWspN}{329}

\newcommand{\nDiffFtHeldA}{$+$0.53}
\newcommand{\nDiffFtHeldC}{$-$0.39}

\newcommand{\nDiffLoraHeldA}{$+$1.13}
\newcommand{\nDiffLoraHeldC}{$+$0.21}

\newcommand{\nDiffPftHeldA}{$-$0.51}
\newcommand{\nDiffPftHeldC}{$-$1.30}

\newcommand{\nDiffWLppmlDevA}{$+$1.31}
\newcommand{\nDiffWLppmlDevB}{$+$0.98}
\newcommand{\nDiffWLppmlDevC}{$+$0.92}
\newcommand{\nDiffWLppmlHeldA}{$+$3.37}
\newcommand{\nDiffWLppmlHeldC}{$+$1.67}
\newcommand{\nFtDevA}{93.22}

\newcommand{\nLoraDevA}{93.42}

\newcommand{\nLpDevA}{87.21}

\newcommand{\nLppDevA}{90.09}

\newcommand{\nLppHeldA}{68.83}
\newcommand{\nLppHeldC}{72.02}
\newcommand{\nLppmlDevA}{92.46}
\newcommand{\nLppmlDevB}{93.31}
\newcommand{\nLppmlDevC}{94.07}
\newcommand{\nLppmlHeldA}{67.37}
\newcommand{\nLppmlHeldC}{72.22}
\newcommand{\nNiFtAAhead}{8 of 11}

\newcommand{\nNiHeldLoraAP}{0.052}
\newcommand{\nNiHeldLoraCP}{0.064}

\newcommand{\nNiLoraAAhead}{7 of 11}

\newcommand{\nPftDevA}{93.85}
\newcommand{\nPftDevB}{94.29}

\newcommand{\nWspDevA}{93.77}
\newcommand{\nWspDevB}{94.29}

\newcommand{\nZsDevA}{87.44}

\newcommand{\nLoWspA}{91.30}
\newcommand{\nLoFtA}{87.82}
\newcommand{\nLoLoraA}{88.35}
\newcommand{\nLoPftA}{91.91}
\newcommand{\nLoZsA}{87.25}
\newcommand{\nLoLpA}{77.32}
\newcommand{\nLoLppA}{86.48}
\newcommand{\nLoDFtA}{$+$3.48}
\newcommand{\nLoDLoraA}{$+$2.94}
\newcommand{\nLoDPftA}{$-$0.61}
\newcommand{\nLoWspB}{92.37}
\newcommand{\nLoFtB}{90.69}
\newcommand{\nLoLoraB}{91.20}
\newcommand{\nLoPftB}{92.84}

\newcommand{\nLoDFtB}{$+$1.68}
\newcommand{\nLoDLoraB}{$+$1.18}
\newcommand{\nLoDPftB}{$-$0.47}

\newcommand{\nFsWspDev}{90.03}
\newcommand{\nFsFtDev}{84.71}
\newcommand{\nFsLoraDev}{86.24}
\newcommand{\nFsPftDev}{90.50}
\newcommand{\nFsDFtDev}{$+$5.31}

\newcommand{\nFsDLoraDev}{$+$3.79}

\newcommand{\nFsDPftDev}{$-$0.47}
\newcommand{\nFsCIPftDev}{$[-1.85, +0.92]$}
\newcommand{\nFsAheadPftDev}{6 of 11}
\newcommand{\nFsWspHeld}{63.81}
\newcommand{\nFsFtHeld}{61.44}
\newcommand{\nFsLoraHeld}{61.29}
\newcommand{\nFsPftHeld}{65.20}
\newcommand{\nFsDFtHeld}{$+$2.37}

\newcommand{\nFsDLoraHeld}{$+$2.52}

\newcommand{\nFsDPftHeld}{$-$1.39}

\definecolor{rowgray}{gray}{0.94}          %
\definecolor{wspblue}{RGB}{226,236,250}    %
\newcolumntype{W}{>{\columncolor{wspblue}}r}
\newcolumntype{V}{>{\columncolor{wspblue}}c}   %
\newcommand{\wsp}{\textsc{WSP}}
\newcommand{\mask}{\texttt{<mask>}}

\title{Fine-Tuning Without Gradients?\\Closed-Form, LLM-Written Classifiers on a Frozen Encoder}

\author{Md Nasir Uddin\textsuperscript{1}\thanks{Research Associate, CSE, BUET. Funded by BIRDI Grant EOI122 under SICIP Project.} \quad
  Musa Tur Farazi\textsuperscript{1} \quad
  Soumik Bhattacharjee\textsuperscript{1} \\
  \textbf{Tishna Sabrina\textsuperscript{2}} \quad
  \textbf{Chowdhury Mohammad Rakin Haider\textsuperscript{1}} \\
  \textsuperscript{1}Department of Computer Science and Engineering, \\
  Bangladesh University of Engineering and Technology (BUET), Dhaka-1000, Bangladesh \\
  \textsuperscript{2}Department of Electrical and Electronic Engineering, \\
  University of Asia Pacific, Dhaka, Bangladesh \\
  \texttt{nasir.iit.du@gmail.com}}

\begin{document}
\maketitle

\begin{abstract}
Can a classifier computed in closed form on a \emph{frozen} encoder match gradient fine-tuning? We present closed-form weight-space programming (\wsp{}): a large language model (LLM), used as a design assistant, writes programs that read a frozen masked language model through cloze-style prompt templates at several layers, fit closed-form heads (shrinkage discriminant analysis, kernel ridge regression and the encoder's label-word prior) on labelled data, and select among a pool of candidate heads by cross-fitting. No gradient is computed and no encoder weight changes. We instantiate the framework on RoBERTa-large. On 11 development datasets with 512 to 8,192 training labels, \wsp{} is ahead of full fine-tuning (FT) and LoRA at every label budget (by 0.3 to 0.6 points on average; the registered non-inferiority tests pass) and within 0.5 points of prompt-based fine-tuning (PFT); these datasets also shaped the recipe. With fewer labels its lead over FT and LoRA grows: 1.2 to 3.5 points with 64 and 128 labels, and 3.8 to 5.3 points in the few-shot setting of 16 labels per class, where \wsp{} is again within half a point of PFT on the development datasets. On six held-out datasets, \wsp{} stays ahead of LoRA and within half a point of FT on average, although with six datasets no registered test passes, and PFT is ahead, by 1.30 points with 8,192 labels. \wsp{} never collapses and is deterministic, but it needs one encoder pass per template (17 here). We report all eight recipe versions, every decision that saw test results, and 27 ideas that did not help.
\end{abstract}

\section{Introduction}
Fine-tuning dominates task adaptation: full fine-tuning (FT), parameter-efficient variants such as LoRA \citep{hu2022lora}, and, with few labels, prompt-based fine-tuning \citep[PFT;][]{gao2021making}. They need gradient training per task, a learning-rate search, a dev set to select on, and a copy of the weights or an adapter per task, and fine-tuning large encoders on small data can collapse to a constant classifier \citep{dodge2020fine,mosbach2021stability}. A frozen encoder already holds much of what a few-class task needs: linear and prompted probes of frozen representations are long-standing baselines \citep{peters2019tune,conneau2018senteval,cho2023prompt}, and gradient-free classifiers on frozen masked language models read through prompts have been compared with fine-tuning before \citep{sun2022bbt,sun2022bbtv2,hou2023promptboosting}. In vision, training-free adaptation of frozen CLIP-style encoders mixes a zero-shot classifier with closed-form heads \citep{zhang2022tip,wang2024hard}. We ask how far a closed-form readout of a frozen text encoder can go when the readout itself is a selected program over many prompts, layers and heads, and where the remaining gap to fine-tuning lies.

\paragraph{The method, and its name.} We call the method closed-form weight-space programming (\wsp{}); in established terms it is prompt-based probing with closed-form heads and LLM-written prompts. The name follows \citet{takehi2026sparks}, who use \emph{weight-space programming} for an LLM that reads and writes a network's weights directly, and \citet{ok2026weightspace}, who let LLMs write the weights of small models by hand. Our variant departs from that meaning: here the LLM writes no weight. It writes a \emph{program} (which cloze templates to wrap an input in, which layers and token positions to read, which closed-form heads to fit and how to select among them), and the classifier's weights are then computed from labelled data by classical estimators: shrinkage linear discriminant analysis (LDA), exact kernel ridge regression and the encoder's own masked-language-model (MLM) label-word prior. The LLM's role is design assistance; we did not compare its programs with human-written or published templates, so the paper makes no claim that LLM authorship itself helps (\S\ref{sec:method}, Appendix~\ref{app:editor}).

\paragraph{When would one choose it?} \wsp{} is meant for a setting in which one frozen encoder serves many classification tasks and per-task gradient training is unwanted or unavailable: for example a shared inference service that must add tasks without training jobs, or a deployment that needs bit-for-bit reproducible classifiers whose adaptation step is a closed-form fit. Its costs fall elsewhere. Whoever serves the classifier pays one encoder pass per template instead of one pass in all (17 in our instantiation), and kernel heads keep the training features of every view (\S\ref{sec:cost}); whoever builds a task still needs the same labelled training and dev data as fine-tuning, and the design of the recipe used LLM rounds. When one can afford a large LLM per prediction, classifying with it directly may be simpler; when one can afford training, PFT remains the stronger choice on new tasks in our study.

\paragraph{How the evidence should be read.} Every method sees the same labelled subsets (paired by seed) and the same dev set, used only for selection; the trained baselines get a learning-rate grid, step-matched training and the best checkpoint on dev. Every \wsp{} recipe version and every statistical test was written down with a time stamp before its own results were computed, and every version is reported. This is an internal, time-stamped plan, not a public pre-registration, and it did not stop the development test sets from informing the design: some decisions were taken with their test results in view (Appendix~\ref{app:prereg} lists all of them). The results on the 11 development datasets are therefore development results. Six held-out datasets, run with fixed programs, are the cleaner check, but with six datasets that check has little power (\S\ref{sec:heldout}).

\paragraph{Findings.}
\begin{itemize}
\item \emph{Better than FT and LoRA} (\S\ref{sec:main}, \S\ref{sec:lowlabel}, \S\ref{sec:fewshot}). On the 11 development datasets, \wsp{}'s averages are above FT and LoRA at every label budget, and the registered non-inferiority tests pass. With 64 and 128 labels the lead grows to 1.2 to 3.5 points, and with 16 labels per class and a dev set of the same size to 3.8 (LoRA) and 5.3 (FT) points (\S\ref{sec:fewshot}). On the six held-out datasets, \wsp{} is ahead of LoRA and within about half a point of FT on average (ahead of FT with 512 labels, behind with 8,192), though six datasets are too few for the tests to pass.
\item \emph{Competitive with PFT on the development datasets, behind it on new tasks} (\S\ref{sec:tests}, \S\ref{sec:heldout}). On the development datasets \wsp{} is within the 0.5-point margin of PFT, which reads one template where \wsp{} reads eight (and 16 for its prior); with 64 and 128 labels it is 0.61 and 0.47 points behind, and with 16 labels per class 0.47 points behind, while ahead on \nFsAheadPftDev{} datasets. On the held-out datasets PFT is ahead by 0.51 points with 512 labels and by 1.30 with 8,192, the latter beyond the margin (90\% interval $-$1.96 to $-$0.64), and by 1.39 with 16 labels per class, and every trained baseline is ahead on CoLA with 512 or more labels.
\item \emph{Probes} (\S\ref{sec:probes}). On the same frozen encoder, \wsp{} is above every probe on every dataset at every budget, development and held-out, with one exception (LP-prompt on Yahoo-10 with 512 labels). This includes a probe that reads its 16 templates in 12 layers, by 0.9 to 1.3 points on average on the development datasets and 1.7 to 3.4 on the held-out ones. With those views, a logistic probe comes within about one point of FT and LoRA on the development datasets, but stays 1.5 to 2.8 points behind them on the held-out ones.
\item \emph{Stability and cost} (\S\ref{sec:cost}). \wsp{} never collapsed and is deterministic for a fixed program, but it reads one encoder pass per template (17 here), takes 2.3 to 4.8 times as long to adapt as a tuned baseline (SST-2), and its kernel heads store gigabytes at 8,192 labels.
\item \emph{What mattered} (\S\ref{sec:ladder}). The steps that moved \wsp{} changed how heads are selected and how the task is phrased in the prompt; 27 other ideas, including closed-form edits of the encoder's weights and demonstrations in the input, did not help.
\end{itemize}

\noindent Our main lesson is where the gap between a frozen encoder and fine-tuning lies on these tasks. Against FT and LoRA, most of it is a readout gap: on the development datasets a logistic probe on many frozen views comes within about one point, and \wsp{}'s heads and selection cover the rest; on the held-out datasets the same probe is 1.5 to 2.8 points behind, while \wsp{} is within about half a point of FT and ahead of LoRA on average. Against PFT on new tasks, a gap of about one point remains that no frozen-side idea we tried closed; we read it, as a hypothesis, as the part of fine-tuning that adapts the representation itself.

\section{Method}
\label{sec:method}
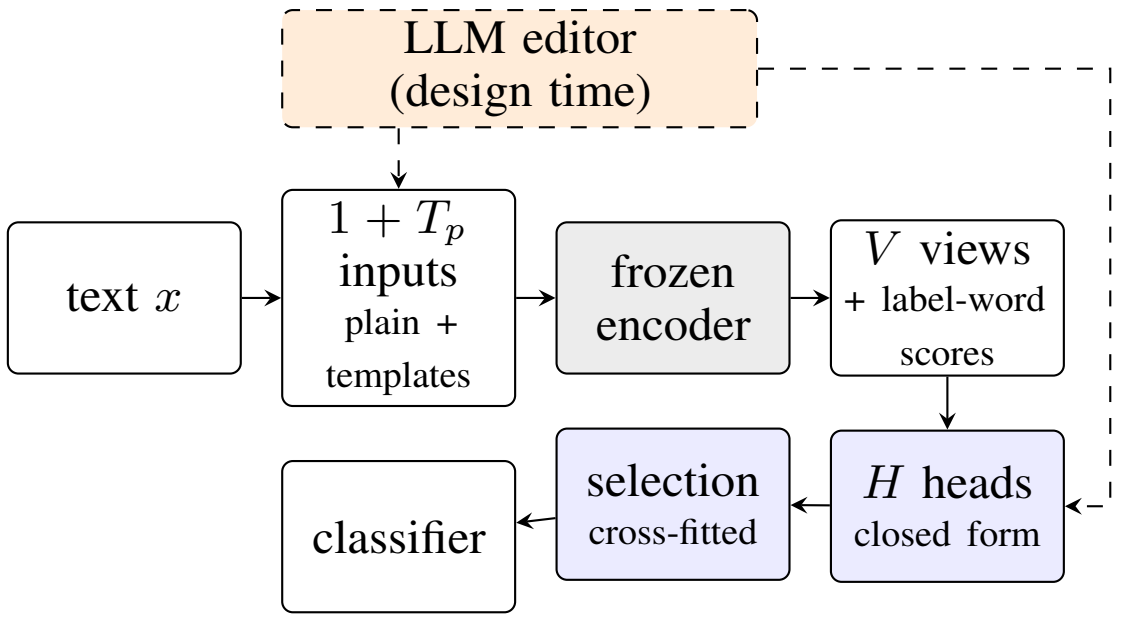
\begin{figure}[t]
\centering
\begin{tikzpicture}[
  font=\footnotesize, >=stealth, node distance=3.5mm and 2.6mm,
  box/.style={draw, rounded corners=2pt, align=center, inner sep=2pt, minimum height=10mm, text width=14mm},
  frozen/.style={box, fill=gray!15},
  closed/.style={box, fill=blue!8},
  edit/.style={box, fill=orange!15, dashed, text width=30mm, minimum height=6mm}]
\node[box] (x) {text $x$};
\node[box, right=of x] (tpl) {$1+T_p$ inputs\\[-1pt]{\scriptsize plain + templates}};
\node[frozen, right=of tpl] (enc) {frozen\\ encoder};
\node[box, right=of enc] (views) {$V$ views\\[-1pt]{\scriptsize + label-word scores}};
\node[closed, below=of views] (heads) {$H$ heads\\[-1pt]{\scriptsize closed form}};
\node[closed, below=of enc] (sel) {selection\\[-1pt]{\scriptsize cross-fitted}};
\node[box, below=of tpl] (clf) {classifier};
\node[edit, above=4mm of tpl, xshift=8mm] (ed) {LLM editor (design time)};
\draw[->] (x) -- (tpl); \draw[->] (tpl) -- (enc); \draw[->] (enc) -- (views); \draw[->] (views) -- (heads);
\draw[->] (heads) -- (sel); \draw[->] (sel) -- (clf);
\draw[->, dashed] (ed.south -| tpl) -- (tpl);
\draw[->, dashed] (ed.east) -| ([xshift=3mm]views.east) |- (heads.east);
\end{tikzpicture}
\caption{Closed-form weight-space programming. At design time an LLM editor writes the program: the cloze templates a text is wrapped in, the views read from the frozen encoder, the closed-form heads (shrinkage LDA, kernel ridge, the MLM label-word prior) and the rule that selects among them. The plain input and $T_v$ view templates are read as $V=(1+T_v)\times 2\times L$ views (2 poolings at each of $L$ layers); $T_p \ge T_v$ templates give label-word scores at \mask{}, so one prediction reads the encoder $1+T_p$ times. In our RoBERTa-large instantiation, $T_v=8$, $T_p=16$, $L=12$, $V=216$ and $H=113$. The classifier's weights are computed from labelled data, not trained, and no encoder weight changes.}
\label{fig:overview}
\end{figure}

\paragraph{Views.} Let $f$ be a frozen masked language model, and let a \emph{template} wrap a text $x$ in a cloze sentence with one \mask{} slot, such as ``\textit{x} It was \mask{} .''. For each text, $f$ is run on the plain input and on each template. A \emph{view} is one pooled hidden state of one input at one layer: for the plain input, the first-token state and the mean of the token states; for a template, the \mask{} state and the mean of the token states. With $T_v$ view templates read at the top $L$ layers, a text gives $V = (1+T_v) \times 2 \times L$ views. Separately, the MLM head's scores for the task's label words at \mask{} are read for a set of $T_p \ge T_v$ templates, which gives the label-word prior; one prediction therefore runs $f$ on $1+T_p$ inputs. The program fixes the templates, the label words and how view templates are chosen; \S\ref{sec:setup} gives the values of our RoBERTa-large instantiation.

\paragraph{Closed-form heads.} Every head is solved exactly (Appendix~\ref{app:heads}, Table~\ref{tab:heads} with every grid): per-view shrinkage LDA \citep{friedman1989regularized,ledoit2004well}, averaged over views; LDA on concatenated per-view principal-component (PCA) projections; a stacked LDA on cross-fitted per-view posteriors; template-pooled LDA; random-Fourier-feature ridge \citep{rahimi2007random}; exact linear, radial-basis-function (RBF) and additive multi-view RBF kernel ridge in dual form; and the zero-shot label-word prior of the frozen MLM head \citep[calibrated as in][]{zhao2021calibrate}, alone, as LDA features, as an additive mixture with a supervised head, or as the prior mean of a kernel ridge. With their hyperparameter grids, the families give a pool of $H$ candidate heads per run.

\paragraph{Selection.} The selection rule never reads the test set. Its held-out rows are the dev set plus 5-fold cross-fitted predictions on the labelled training set \citep{stone1974cross} (with our budgets, 1.5 to 9 times as many rows as the dev set). Three rules are candidates: the best head, the average of the top three, and Occam's window \citep[every head within one standard error of the best, the one-standard-error rule of][]{breiman1984cart,madigan1994model}; the rule is itself chosen by 20 random half-splits of the held-out rows. The final heads are fitted on all labelled training data.

\paragraph{The editor.} The editor is an LLM; in this study, Claude (Anthropic), used through the Claude Code agent (Appendix~\ref{app:editor}). It proposed every program family, template, label-word set, selection rule and registered round, implemented and ran them, and drafted the analyses; the authors set the goals and constraints, left the technical decisions to the agent, and checked the results. When writing or revising a program the editor reads dev data, never test data; but as the agent that ran the study it saw the test results of every earlier version, one more reason why the development results are development results. Its rounds are of two kinds: \emph{editor rounds}, which rewrite the templates of a task and adopt the rewrite only if it gains at least one point of \mask{}-LDA dev accuracy, and \emph{registered rounds}, each one idea with a pass rule fixed before its first run, scored on dev data and, from the demonstrations round on, on dev and cross-fitted train rows of a fresh data subset (\S\ref{sec:ladder}).

\section{Experimental setup}
\label{sec:setup}
\paragraph{Datasets.} Eleven development datasets: SST-2 \citep{socher2013recursive}, AG News, Yelp-2 and Amazon-2 \citep{zhang2015character}, MR \citep{pang2005seeing}, SUBJ \citep{pang2004sentimental}, CR \citep{hu2004mining}, MPQA \citep{wiebe2005annotating}, TREC-6 \citep{li2002learning}, BBC News \citep{greene2006practical} and IMDB \citep{maas2011learning}. Six held-out datasets: SST-5, CoLA \citep{warstadt2019neural}, DBpedia-14, Yahoo-10, Yelp-5 and Amazon-5 \citep{zhang2015character}. All are English single-text tasks with 2 to 14 classes and label words of one token, which is what we mean by \emph{few-class}. Eight of the 11 development datasets are binary, against one of the six held-out datasets. Details are in Appendix~\ref{app:setup}.

\paragraph{Encoder and recipe.} We instantiate \wsp{} on RoBERTa-large \citep{liu2019roberta} (24 layers, 1,024 dimensions). The final recipe (v8) reads the top $L=12$ layers and $T_v=8$ view templates: the four of the task's program and four of a second batch of twelve (written by the editor before any of them was scored), chosen per data subset by 5-fold train-only cross-validated accuracy of the \mask{}-state LDA (Appendix~\ref{app:templates} lists every template and how often each was chosen). That gives $V=(1+8)\times 2\times 12=216$ views of 1,024 dimensions. The prior reads all $T_p=16$ templates, so one prediction reads the encoder 17 times, and the head pool has $H=113$ candidates (Table~\ref{tab:heads}).

\paragraph{Labels.} Each method is run with 512, 2,048 and 8,192 labelled training examples (the whole pool when it is smaller; BBC News has no 2,048-label setting), on 3 data subsets drawn with seeds 0--2; a registered extension adds 64 and 128 labels on the development datasets (\S\ref{sec:lowlabel}), and a second the few-shot setting of LM-BFF, 16 labels per class with a dev set of 16 per class, on all 17 datasets (\S\ref{sec:fewshot}). A fixed dev split of 1,024 examples (BBC News: 245) is carved from train and used by every method for selection, so the total labelled data is 1,536, 3,072 and 9,216 examples. This is far more than the 16 examples per class for which PFT was designed \citep{gao2021making}; our budgets test it outside its few-shot regime.

\paragraph{Baselines.} FT, LoRA \citep[rank 8 on query and value;][]{hu2022lora} and PFT \citep[as LM-BFF without demonstrations;][]{gao2021making}, all on the same subsets, with seed $k$ trained on subset $k$. Each is trained for 768 optimizer steps (step-matched across budgets; 654 to 834 where the whole, smaller pool is used) with two learning rates, and the learning rate and checkpoint are chosen on dev for each seed (Appendix~\ref{app:setup} gives every setting and the learning rates selected). PFT trains every weight through the \emph{first} template and the label words of the task's \wsp{} program. It therefore shares \wsp{}'s label words and one of its templates, but not the other seven view templates nor the 16-template prior: the comparison with PFT confounds two contrasts: closed-form readout against gradient training, and many prompts against one. Probes on the same frozen encoder: LP, logistic regression on the last layer of the plain input, and LP-prompt, the same on the \mask{} state of PFT's template, with the regularisation chosen on dev; two multi-layer probes, one of them over 16 templates (\S\ref{sec:probes}), read many of the views \wsp{} reads.

\paragraph{Statistics.} Cells are means over the 3 subsets or seeds; $\pm$ gives the standard deviation with divisor 3, a descriptive spread of the three runs (the sample standard deviation, divisor 2, is $\sqrt{3/2} \approx 1.22$ times larger); the tests below use the per-dataset means. Across datasets, we test non-inferiority at a margin of 0.5 points: a one-sided paired $t$-test of $H_0$: mean(\wsp{} $-$ baseline) $\le -0.5$, together with the 5th percentile of 10,000 bootstrap resamples of datasets; both must clear the margin. The margin was fixed after the 11-dataset differences against FT and LoRA were known and before any PFT or held-out result; it is about the size of the spread of a method across data subsets on our smaller test sets (Appendix~\ref{app:tables}). We report, for every comparison, the two-sided 90\% $t$-interval and the complementary test of inferiority beyond the margin ($H_0$: mean $\ge -0.5$), and we make no multiplicity correction across the 15 registered comparisons. With a Bonferroni correction ($\alpha = 0.05/15$), every development test against FT and LoRA would still pass (the closest, LoRA at 512 labels, has $p=0.0028$), but against PFT only the one at 8,192 labels would ($p=0.031$ and $0.006$ at 512 and 2,048 labels). A baseline run counts as collapsed when its best dev accuracy is within one point of the majority-class rate; collapsed seeds are left out of the baseline means (this only changes LoRA on MPQA at 8,192 labels).

\section{Results}
\label{sec:results}
\begin{table*}[t]
\centering\small
\setlength{\tabcolsep}{2.4pt}
\begin{tabular}{lrcccccVcccc}
\toprule
& & \multicolumn{6}{c}{\textit{frozen encoder, no gradient training}} & \multicolumn{3}{c}{\textit{gradient training}} & \\
\cmidrule(lr){3-8}\cmidrule(lr){9-11}
Datasets & Labels & Zero-shot & LP & LP-prompt & LP-ml & \shortstack{LP-prompt\\-ml} & \textbf{WSP (ours)} & FT & LoRA & PFT & WSP $-$ PFT \\
\midrule
11 development & 512 & 87.44 & 87.21 & 90.09 & 89.73 & 92.46 & \underline{93.77} & 93.22 & 93.42 & \textbf{93.85} & $-$0.08 \\
10 development & 2,048 & 86.97 & 89.38 & 91.11 & 91.61 & 93.31 & \textbf{94.29} & 93.95 & 93.87 & \textbf{94.29} & $\pm$0.00 \\
11 development & 8,192 & 87.56 & 91.32 & 92.54 & 93.03 & 94.07 & \underline{94.99} & 94.65 & 94.63 & \textbf{95.08} & $-$0.09 \\
\midrule
6 held-out & 512 & 59.12 & 63.00 & 68.83 & 63.49 & 67.37 & \underline{70.74} & 70.21 & 69.61 & \textbf{71.26} & $-$0.51 \\
6 held-out & 8,192 & 59.21 & 69.52 & 72.02 & 70.27 & 72.22 & 73.89 & \underline{74.28} & 73.68 & \textbf{75.19} & $-$1.30 \\
\bottomrule
\end{tabular}
\caption{Average test accuracy (\%) over datasets. Every method is trained or fitted on the same labelled examples and selects on the same 1,024-example dev set. \textbf{Bold}: the best of WSP, FT, LoRA and PFT in the row; \underline{Underlined}: the second best (none when two tie for the best); many gaps are a few hundredths of a point, so the ranks are descriptive, and the registered tests (Table~\ref{tab:ni}) are the evidence. Zero-shot: the calibrated label-word scores of the frozen MLM head; it fits no head, but its calibration uses the training subset and, for the datasets revised by the editor rounds, its prompts were chosen on dev, so it varies slightly with the budget. LP and LP-prompt are linear probes on the last layer of the same frozen encoder (LP-prompt: PFT's template). LP-ml and LP-prompt-ml read many views, as WSP does: the top 12 layers of the plain input with both poolings (24 views), and the \mask{} state of the 16 templates of WSP's prior in the top 12 layers (192 views), each view reduced to principal components (6,144 features). WSP: mean over 3 data subsets (one run where the whole pool is used); FT, LoRA, PFT: mean over 3 seeds on the same subsets (collapsed seeds left out). WSP $-$ PFT: the mean of the per-dataset differences, from unrounded accuracies. Development rows: these datasets shaped the recipe; held-out rows: programs fixed before any run on them. BBC News has no 2,048-label row (its pool is smaller).}
\label{tab:headline}
\end{table*}

\begin{table*}[t]
\centering\small
\setlength{\tabcolsep}{6pt}
\rowcolors{1}{white}{rowgray}
\begin{tabular}{lrrrrrrrr}
\toprule
\hiderowcolors
& \multicolumn{4}{c}{\textbf{512 labels}} & \multicolumn{4}{c}{\textbf{8,192 labels} (whole pool when smaller)} \\
\cmidrule(lr){2-5}\cmidrule(lr){6-9}
Dataset & \textbf{WSP} & FT & LoRA & PFT & \textbf{WSP} & FT & LoRA & PFT \\
\midrule
\showrowcolors
SST-2 & \underline{93.65} & 92.85 & 92.97 & \textbf{94.00} & \underline{95.11} & 94.15 & 94.30 & \textbf{95.18} \\
AG News & \underline{90.10} & \textbf{90.11} & 89.63 & 89.92 & \underline{93.11} & 92.93 & 92.36 & \textbf{93.23} \\
MR & \underline{90.68} & 88.81 & 88.49 & \textbf{90.81} & 91.18 & \underline{91.34} & \underline{91.34} & \textbf{91.84} \\
SUBJ & \underline{96.38} & 95.50 & 96.20 & \textbf{96.62} & \textbf{97.65} & 97.38 & 97.27 & \textbf{97.65} \\
CR & 92.73 & 92.55 & \underline{92.82} & \textbf{93.62} & \underline{94.15} & 93.35 & 92.82 & \textbf{94.24} \\
MPQA & 87.37 & 87.40 & \underline{88.28} & \textbf{88.68} & \underline{90.40} & 89.90 & 89.70 & \textbf{90.52} \\
TREC-6 & \underline{96.20} & 95.27 & \textbf{96.60} & 95.47 & \textbf{97.20} & 96.60 & \textbf{97.20} & 97.00 \\
BBC News & 97.47 & \underline{97.50} & 97.47 & \textbf{97.90} & 97.90 & 97.67 & \textbf{98.03} & \underline{97.93} \\
Yelp-2 & \textbf{96.85} & 96.71 & \underline{96.73} & 96.68 & \underline{97.36} & 97.29 & 97.34 & \textbf{97.55} \\
Amazon-2 & \textbf{96.15} & \underline{95.51} & 95.43 & 95.49 & \textbf{96.37} & \underline{96.29} & 96.23 & 96.15 \\
IMDB & \textbf{93.91} & \underline{93.18} & 92.97 & \underline{93.18} & \underline{94.49} & 94.27 & 94.35 & \textbf{94.57} \\
\midrule
\rowcolor{white}
\textit{Average} & \underline{93.77} & 93.22 & 93.42 & \textbf{93.85} & \underline{94.99} & 94.65 & 94.63 & \textbf{95.08} \\
\bottomrule
\end{tabular}
\caption{Test accuracy (\%) on each development dataset. \textbf{Bold}: the best of the four methods at that label budget; \underline{Underlined}: the second best (none when two tie for the best); many gaps are a few hundredths of a point, so the ranks are descriptive, and the registered tests (Table~\ref{tab:ni}) are the evidence. Means over 3 data subsets (WSP; one run where the whole pool is used) or 3 seeds (baselines); standard deviations and the 2,048-label results are in Appendix~\ref{app:tables}. }
\label{tab:pertask}
\end{table*}

\subsection{Development datasets}
\label{sec:main}
Table~\ref{tab:headline} gives the averages and Table~\ref{tab:pertask} the datasets; these are development results. With 512 labels \wsp{} averages \nWspDevA\%, against \nFtDevA\% for FT, \nLoraDevA\% for LoRA and \nPftDevA\% for PFT; with 2,048 labels it is level with PFT (\nWspDevB\% against \nPftDevB\%) and ahead of FT and LoRA. The margins are a few tenths of a point, and per dataset they are not uniform: at 512 labels \wsp{} is above FT on \nNiFtAAhead{} datasets and above LoRA on \nNiLoraAAhead{}, with one tie (sign tests, ties left out: $p=0.23$ and $0.34$); the evidence for the average lead comes from the paired tests below, not from a consistent sign.

\paragraph{Where \wsp{} wins and loses (512 labels).} Against FT and LoRA, \wsp{} wins most on sentence and review sentiment (MR, IMDB, SST-2, Amazon-2) and loses most on short opinion phrases (MPQA). Against PFT the picture is mixed: \wsp{} is ahead on the long reviews (IMDB, Amazon-2, Yelp-2), on TREC-6 and on AG News, and behind on the short opinion and review texts, most of all MPQA and CR, where PFT leads by about one point, and on BBC News. Our hypothesis is that fine-tuning through a prompt adapts the representation, which a closed-form readout of a frozen encoder cannot; we did not test it directly.

\paragraph{What the program selects.} The selection trace of every run is saved. In the runs with 512 labels, the rule usually averages many heads (Occam's window in two thirds of the runs), and the best single head is most often a supervised head mixed with the zero-shot prior: with few labels, the encoder's own label-word prior carries real weight. With more labels the additive multi-view kernel takes over and fewer heads are averaged. No family is best everywhere.

\begin{table}[t]
\centering\small
\setlength{\tabcolsep}{2.4pt}
\begin{tabular}{llccc}
\toprule
& & \multicolumn{3}{c}{WSP $-$ baseline (points)} \\
\cmidrule(lr){3-5}
Data & Labels & vs FT & vs LoRA & vs PFT \\
\midrule
Dev. & 512 & \textbf{$+$0.56}\;\checkmark & $+$0.35\;\checkmark & $-$0.08\;\checkmark \\
 & & {\scriptsize\color{black}[$+$0.24, $+$0.87]} & {\scriptsize\color{black}[$-$0.09, $+$0.80]} & {\scriptsize\color{black}[$-$0.44, $+$0.28]} \\[1pt]
 & 2,048 & \textbf{$+$0.34}\;\checkmark & \textbf{$+$0.42}\;\checkmark & $\pm$0.00\;\checkmark \\
 & & {\scriptsize\color{black}[$+$0.08, $+$0.59]} & {\scriptsize\color{black}[$+$0.18, $+$0.65]} & {\scriptsize\color{black}[$-$0.29, $+$0.28]} \\[1pt]
 & 8,192 & \textbf{$+$0.34}\;\checkmark & \textbf{$+$0.36}\;\checkmark & $-$0.09\;\checkmark \\
 & & {\scriptsize\color{black}[$+$0.16, $+$0.53]} & {\scriptsize\color{black}[$+$0.10, $+$0.62]} & {\scriptsize\color{black}[$-$0.21, $+$0.04]} \\[1pt]
\midrule
Held-out & 512 & $+$0.53\;$\times$ & $+$1.13\;$\times$ & $-$0.51\;$\times$ \\
 & & {\scriptsize\color{black}[$-$0.74, $+$1.80]} & {\scriptsize\color{black}[$-$0.52, $+$2.79]} & {\scriptsize\color{black}[$-$1.29, $+$0.26]} \\[1pt]
 & 8,192 & $-$0.39\;$\times$ & $+$0.21\;$\times$ & $-$1.30\;$\ddagger$ \\
 & & {\scriptsize\color{black}[$-$1.33, $+$0.54]} & {\scriptsize\color{black}[$-$0.58, $+$1.00]} & {\scriptsize\color{black}[$-$1.96, $-$0.64]} \\[1pt]
\bottomrule
\end{tabular}
\caption{Registered tests across datasets. Each cell: the mean of WSP minus the baseline (points, from unrounded accuracies, so it can differ by 0.01 from the difference of the rounded averages in Table~\ref{tab:headline}), and below it the two-sided 90\% $t$-interval; \checkmark: non-inferior at a 0.5-point margin; $\times$: not shown; $\ddagger$: inferior by more than the margin (the whole interval below $-0.5$); \textbf{bold}: WSP ahead with 95\% confidence (bootstrap lower bound above 0). Development rows are development results (these datasets shaped the recipe). Bounds and $p$-values: Table~\ref{tab:app-ni}.}
\label{tab:ni}
\end{table}

\subsection{Registered tests}
\label{sec:tests}
Table~\ref{tab:ni} gives the registered tests with their 90\% intervals. On the development datasets, \wsp{} is non-inferior to FT, LoRA and PFT at every budget; against PFT the intervals are $[-0.44, +0.28]$, $[-0.29, +0.28]$ and $[-0.21, +0.04]$ at 512, 2,048 and 8,192 labels, so the margin is cleared by 0.06 to 0.29 points, a distance of the same order as the steps of the recipe ladder that the development datasets informed. Against FT the intervals lie above zero at every budget, and against LoRA at 2,048 and 8,192 labels. On the held-out datasets no test passes. Against LoRA, \wsp{} is ahead on average but the test narrowly misses ($p=\nNiHeldLoraAP$ and $\nNiHeldLoraCP$). Against FT the intervals are wide ($[-0.74, +1.80]$ and $[-1.33, +0.54]$). Against PFT, the interval at 8,192 labels, $[-1.96, -0.64]$, lies wholly below the margin: there \wsp{} is inferior to PFT by more than 0.5 points (complementary test $p=0.029$; behind on 5 of 6 datasets).

\subsection{Against probes on the same encoder}
\label{sec:probes}
A tuned linear probe on the last layer of the frozen encoder is \nLpDevA\% at 512 labels, no better than the zero-shot prior (\nZsDevA\%), and a prompted probe on one template is \nLppDevA\%. \wsp{} is ahead of both on every development dataset at every budget, by 6.6, 4.9 and 3.7 points (LP) and 3.7, 3.2 and 2.5 points (LP-prompt) on average at 512, 2,048 and 8,192 labels; on the held-out datasets LP-prompt is ahead on Yahoo-10 with 512 labels. But both read far less than \wsp{}: one layer, and at most one template. Two stronger probes, registered in the revision, read many of \wsp{}'s views: LP-ml takes the top 12 layers of the plain input with both poolings (24 views), and LP-prompt-ml the \mask{} state of the 16 templates of \wsp{}'s prior, which include its eight view templates, in the top 12 layers (192 views). Together they cover 120 of \wsp{}'s 216 views, all but the token means of the templates. Each view is reduced to principal components fitted on the training subset (6,144 features), followed by the same logistic regression with the regularisation chosen on dev. More views help: LP-prompt-ml averages \nLppmlDevA\%, \nLppmlDevB\% and \nLppmlDevC\% on the development datasets at 512, 2,048 and 8,192 labels, 1.5 to 2.4 points above LP-prompt and 0.6 to 1.0 points below FT and LoRA. \wsp{} is still ahead of both new probes on every development dataset at every budget; its average difference to LP-prompt-ml is \nDiffWLppmlDevA, \nDiffWLppmlDevB{} and \nDiffWLppmlDevC{} points (Table~\ref{tab:app-probes}). On the held-out datasets \wsp{} is ahead of both new probes on all six datasets at both budgets, with an average difference to LP-prompt-ml of \nDiffWLppmlHeldA{} and \nDiffWLppmlHeldC{} points, and there more views do not help the probe: LP-prompt-ml averages \nLppmlHeldA\% and \nLppmlHeldC\% at 512 and 8,192 labels, against \nLppHeldA\% and \nLppHeldC\% for LP-prompt, and is 1.5 to 2.8 points below FT and LoRA. So on the development datasets, reading more views accounts for most of \wsp{}'s lead over the last-layer probes, but not all of it: with \wsp{}'s templates and layers, a tuned logistic regression stays about one point behind. On the held-out datasets, a logistic probe gains little or nothing from the same views, and \wsp{}'s lead over the probes stays. We read this as an interpretation, not a decomposition: \wsp{} also differs from these probes in its heads, the label-word prior among them, and in its selection, and we have not separated the two.

\subsection{Held-out datasets}
\label{sec:heldout}
The recipe was built on the development datasets, and some decisions saw their test results (Appendix~\ref{app:prereg}). The held-out datasets were run with programs fixed before any run on them: their first templates were written blind, and a mechanical rule fixed beforehand gave five of the six datasets one dev-only editor round. They are a cleaner check, with three limits. (i) \emph{Power}: with six datasets, the registered test passes a truly level method only if the standard deviation of the per-dataset differences is below about 0.61 points; the observed values are 0.80 to 2.01. (ii) \emph{Task mix}: five of the six held-out datasets have 5 to 14 classes, against two of the 11 development datasets. (iii) \emph{Sequencing}: the second batch (Yahoo-10, Yelp-5, Amazon-5) was run after the first batch's results were known, the held-out datasets were scored with v7 before v8, and the batch of templates that became v8's extra views was written for every task, held-out ones included.

There, the average difference between \wsp{} and FT is \nDiffFtHeldA{} points with 512 labels and \nDiffFtHeldC{} with 8,192; against LoRA it is \nDiffLoraHeldA{} and \nDiffLoraHeldC{}, and against PFT \nDiffPftHeldA{} and \nDiffPftHeldC{} (Table~\ref{tab:headline}, Appendix~\ref{app:tables}). \wsp{} is the best method on fine-grained SST-5 with 512 labels and level with all three on DBpedia-14, but it is behind every trained baseline on CoLA (Matthews correlation in Appendix~\ref{app:tables}), where the zero-shot prior of the frozen model is below the majority rate (45.1\% against 69.1\%). PFT's lead grows with more labels, and v8's extra templates bring no gain over v7 on these datasets ($-$0.13 and $+$0.05 points).

\begin{table*}[t]
\centering\small
\setlength{\tabcolsep}{6pt}
\rowcolors{3}{white}{rowgray}
\begin{tabular}{lrrrrrrrr}
\toprule
& \multicolumn{4}{c}{\textbf{64 labels}} & \multicolumn{4}{c}{\textbf{128 labels}} \\
\cmidrule(lr){2-5}\cmidrule(lr){6-9}
Dataset & \textbf{WSP} & FT & LoRA & PFT & \textbf{WSP} & FT & LoRA & PFT \\
\midrule
SST-2 & \textbf{93.08} & 86.09 & 88.53 & \underline{92.43} & \underline{92.70} & 89.95 & 91.82 & \textbf{92.93} \\
AG News & \underline{88.66} & 88.23 & 88.33 & \textbf{89.05} & 88.89 & \underline{89.21} & 89.06 & \textbf{89.43} \\
MR & \textbf{90.43} & 84.43 & 85.87 & \underline{89.62} & \textbf{90.24} & 85.74 & 87.15 & \underline{89.93} \\
SUBJ & \underline{94.37} & 93.60 & 93.58 & \textbf{95.22} & \underline{94.70} & 94.15 & 94.43 & \textbf{95.83} \\
CR & \underline{90.16} & 82.80 & 87.06 & \textbf{92.82} & \underline{91.93} & 87.94 & 90.87 & \textbf{93.17} \\
MPQA & \underline{80.17} & 72.15 & 68.43 & \textbf{84.65} & \underline{84.37} & 80.43 & 78.10 & \textbf{87.90} \\
TREC-6 & \underline{90.20} & 81.53 & 83.93 & \textbf{90.53} & \underline{92.80} & 91.20 & 92.07 & \textbf{93.93} \\
BBC News & 94.67 & \underline{95.90} & 95.50 & \textbf{96.57} & \underline{96.73} & 96.57 & \textbf{97.30} & \underline{96.73} \\
Yelp-2 & 95.43 & \textbf{95.87} & \underline{95.56} & 94.90 & 95.89 & 95.81 & \textbf{95.90} & \textbf{95.90} \\
Amazon-2 & \textbf{94.95} & 94.18 & \underline{94.22} & 93.81 & \textbf{94.95} & 94.77 & \underline{94.81} & 94.74 \\
IMDB & \textbf{92.17} & 91.26 & 90.88 & \underline{91.36} & \textbf{92.90} & \underline{91.84} & 91.65 & 90.71 \\
\midrule
\rowcolor{white}
\textit{Average} (11 / 11) & \underline{91.30} & 87.82 & 88.35 & \textbf{91.91} & \underline{92.37} & 90.69 & 91.20 & \textbf{92.84} \\
\bottomrule
\end{tabular}
\caption{Low-label extension: test accuracy (\%) on each development dataset with 64 and 128 labelled training examples. Every method selects on the same 1,024-example dev set (BBC News: 245), more labelled data than the training set itself. Means over 3 data subsets (WSP) or 3 seeds on the same subsets (FT, LoRA, PFT; collapsed seeds left out). \textbf{Bold}: the best of the four in the row and budget; \underline{underlined}: the second best (none when two tie for the best); the ranks are descriptive, and the registered tests (Table~\ref{tab:ni-lo}) are the evidence. WSP is v8 unchanged, except for a fix for classes absent from a cross-fitting part, which changes TREC-6 only. Standard deviations, the probes per dataset and the fix: Appendix~\ref{app:lowlabel}.}
\label{tab:lowlabel}
\end{table*}

\subsection{Fewer labels: 64 and 128}
\label{sec:lowlabel}
The main study has no budget below 512 labels. A registered extension (Appendix~\ref{app:lowlabel}), fixed before any of its runs, runs \wsp{} v8 unchanged and the three trained baselines under the study's protocol on the 11 development datasets with 64 and 128 labelled training examples (3 subsets; the baselines train for 768 steps and select learning rate and checkpoint on dev). These are the development datasets, but no part of the recipe was chosen at these budgets. The dev set stays the study's 1,024 examples (BBC News: 245), more labelled data than the training set itself; it is the same for every method and used only for selection, so this is a low-label setting, not few-shot learning in the sense of LM-BFF, which \S\ref{sec:fewshot} tests. Table~\ref{tab:lowlabel} gives the datasets and Table~\ref{tab:ni-lo} the registered tests.

With 64 labels \wsp{} averages \nLoWspA\%, against \nLoFtA\% for FT, \nLoLoraA\% for LoRA and \nLoPftA\% for PFT; with 128 labels it averages \nLoWspB\%, against \nLoFtB\%, \nLoLoraB\% and \nLoPftB\%. The tests give the picture of the main study's development results, with larger gaps to FT and LoRA: \wsp{} is non-inferior to both at both budgets, with the whole 90\% interval above zero (\nLoDFtA{} and \nLoDLoraA{} points with 64 labels, \nLoDFtB{} and \nLoDLoraB{} with 128), and ahead of both on 9 of the 11 datasets with 64 labels. FT and LoRA lose most where their seeds vary most (MPQA, TREC-6 and CR; standard deviations up to 12 points with 64 labels); after the learning-rate choice on dev, no baseline seed stays collapsed. Against PFT no test passes: \wsp{} is \nLoDPftA{} and \nLoDPftB{} points behind on average (90\% intervals $[-1.57, +0.36]$ and $[-1.23, +0.30]$), ahead on 5 and 3 of the 11 datasets, and behind most on MPQA (by 4.5 and 3.5 points) and CR; it is ahead on MR, Amazon-2 and IMDB at both budgets. The zero-shot prior alone averages \nLoZsA\% with 64 labels, within a point of FT, and \wsp{} falls below it on MPQA with 64 labels (by 2.7 points) and slightly on SST-2 and MR with 128; the last-layer probes are far behind (LP \nLoLpA\%, LP-prompt \nLoLppA\% with 64 labels).

\begin{table}[t]
\centering\small
\setlength{\tabcolsep}{3pt}
\begin{tabular}{lccc}
\toprule
& \multicolumn{3}{c}{WSP $-$ baseline (points)} \\
\cmidrule(lr){2-4}
Labels & vs FT & vs LoRA & vs PFT \\
\midrule
64 & \textbf{$+$3.48}\;\checkmark & \textbf{$+$2.94}\;\checkmark & $-$0.61\;$\times$ \\
 & {\scriptsize\color{black}[$+$1.36, $+$5.59]} & {\scriptsize\color{black}[$+$0.93, $+$4.96]} & {\scriptsize\color{black}[$-$1.57, $+$0.36]} \\[1pt]
128 & \textbf{$+$1.68}\;\checkmark & \textbf{$+$1.18}\;\checkmark & $-$0.47\;$\times$ \\
 & {\scriptsize\color{black}[$+$0.70, $+$2.66]} & {\scriptsize\color{black}[$+$0.11, $+$2.24]} & {\scriptsize\color{black}[$-$1.23, $+$0.30]} \\[1pt]
\bottomrule
\end{tabular}
\caption{Registered tests of the low-label extension (11 development datasets), as in Table~\ref{tab:ni}: mean of WSP minus the baseline, two-sided 90\% $t$-interval below; \checkmark: non-inferior at the 0.5-point margin (registered $t$-test and bootstrap bound); $\times$: not shown; $\ddagger$: inferior by more than the margin; \textbf{bold}: WSP ahead with 95\% confidence (bootstrap lower bound above 0), as in Table~\ref{tab:ni}. Full numbers: Table~\ref{tab:app-ni-lo}.}
\label{tab:ni-lo}
\end{table}

\begin{table}[t]
\centering\small
\setlength{\tabcolsep}{4pt}
\begin{tabular}{lVccc}
\toprule
Dataset & \textbf{WSP} & FT & LoRA & PFT \\
\midrule
SST-2 & \textbf{92.36} & 77.20 & 82.68 & \underline{89.93} \\
AG News & 86.89 & \underline{87.42} & 87.35 & \textbf{88.62} \\
MR & \textbf{87.97} & 71.91 & 74.41 & \underline{87.52} \\
SUBJ & \underline{91.56} & 90.31 & 91.05 & \textbf{94.42} \\
CR & \underline{88.99} & 79.36 & 82.98 & \textbf{92.55} \\
MPQA & \underline{81.21} & 74.52 & 73.97 & \textbf{85.25} \\
TREC-6 & \textbf{88.32} & 82.52 & \underline{86.16} & 84.52 \\
BBC News & 94.98 & 96.32 & \underline{96.48} & \textbf{96.94} \\
Yelp-2 & \textbf{95.18} & \underline{94.70} & 93.98 & 93.97 \\
Amazon-2 & \textbf{93.68} & 91.22 & 91.67 & \underline{93.16} \\
IMDB & \textbf{89.14} & 86.36 & 87.86 & \underline{88.56} \\
\midrule
\textit{Average} (11) & \underline{90.03} & 84.71 & 86.24 & \textbf{90.50} \\
\midrule
SST-5 & \underline{47.18} & 39.38 & 40.66 & \textbf{48.44} \\
CoLA & 61.19 & \textbf{69.88} & \underline{66.44} & 57.78 \\
DBpedia-14 & 97.42 & 97.88 & \textbf{98.28} & \underline{98.23} \\
Yahoo-10 & \underline{67.70} & 65.96 & 66.46 & \textbf{69.69} \\
Yelp-5 & \underline{57.18} & 50.94 & 51.21 & \textbf{59.16} \\
Amazon-5 & \underline{52.20} & 44.60 & 44.71 & \textbf{57.87} \\
\midrule
\textit{Average} (6) & \underline{63.81} & 61.44 & 61.29 & \textbf{65.20} \\
\bottomrule
\end{tabular}
\caption{Few-shot extension: test accuracy (\%) with 16 labelled examples per class and a dev set of 16 per class, mean over 5 data subsets (WSP) or 5 seeds on the same subsets (FT, LoRA, PFT; trained as in LM-BFF: batch size 8, 1,000 steps). Top: the 11 development datasets; bottom: the 6 held-out datasets. \textbf{Bold}: the best; \underline{underlined}: the second best (none when two tie for the best). Standard deviations and the registered tests: Table~\ref{tab:app-fs}.}
\label{tab:fewshot}
\end{table}

\subsection{Few-shot: 16 labels per class}
\label{sec:fewshot}
The low-label extension keeps the study's large dev set. A second registered extension (Appendix~\ref{app:fewshot}), also fixed before any of its runs, uses the few-shot setting in which PFT was introduced \citep{gao2021making}: 16 labelled training examples per class and a dev set of 16 per class, on all 17 datasets and 5 data subsets. \wsp{} is v8 unchanged; FT, LoRA and PFT are trained as in LM-BFF (batch size 8, 1,000 steps, learning rate and checkpoint chosen on the small dev set). Table~\ref{tab:fewshot} gives every dataset.

On the development datasets \wsp{} averages \nFsWspDev\%, against \nFsFtDev\% for FT, \nFsLoraDev\% for LoRA and \nFsPftDev\% for PFT. It is non-inferior to FT and LoRA, with average differences of \nFsDFtDev{} and \nFsDLoraDev{} points and the whole 90\% interval above zero, the largest leads of the study: with so few labels, FT and LoRA vary strongly across seeds and fall far behind on SST-2, MR and CR, while \wsp{} has no training run to go astray. Against PFT the test does not pass: the average difference is \nFsDPftDev{} points (90\% interval \nFsCIPftDev), ahead on \nFsAheadPftDev{} datasets. On the held-out datasets \wsp{} averages \nFsWspHeld\%, against \nFsFtHeld\%, \nFsLoraHeld\% and \nFsPftHeld\%: ahead of FT and LoRA (differences \nFsDFtHeld{} and \nFsDLoraHeld{} points), though with six datasets no test passes, and behind PFT (difference \nFsDPftHeld{} points), which is ahead on 5 of the 6. CoLA is again the exception for FT and LoRA, which lead \wsp{} there by 5 to 9 points.

\begin{table}[t]
\centering\small
\setlength{\tabcolsep}{4pt}
\begin{tabular}{llWrrr}
\toprule
Trained on & Labels & \textbf{WSP} & FT & LoRA & PFT \\
\midrule
SST-2 & 512 & \textbf{91.44} & \underline{90.57} & 89.33 & 86.82 \\
 & 2,048 & \underline{93.03} & \textbf{93.17} & 92.65 & 89.65 \\
 & 8,192 & 92.45 & \textbf{93.14} & \underline{92.87} & 89.42 \\
\midrule
each of 6 & 512 & 90.17 & \textbf{91.10} & \underline{90.95} & 90.07 \\
\bottomrule
\end{tabular}
\caption{Out-of-domain accuracy (\%) of sentiment models. \textbf{Bold}: the best in the row; \underline{underlined}: the second best. Top: SST-2 models on IMDB and Yelp reviews. Bottom: mean over the six binary sentiment datasets as sources, each tested on review sets from other corpora. WSP is v7 here, the recipe before v8 (v8 reads 16 templates for its prior instead of 4 and adds four view templates).}
\label{tab:ood}
\end{table}

\subsection{Out of domain (recipe v7)}
Table~\ref{tab:ood} tests sentiment models on review sets from other corpora; this analysis was run with v7, the recipe before v8, whose label-word prior reads 4 templates instead of 16 and which lacks v8's four extra view templates. With SST-2 as the source, PFT is 3 to 5 points below \wsp{} on IMDB and Yelp reviews at every budget. A registered follow-up at 512 labels with all six binary sentiment datasets as sources shows that this is not general: over the six sources, \wsp{} is level with PFT and slightly behind FT and LoRA. PFT's deficit comes from the two sources built from Rotten Tomatoes sentences (SST-2 and MR). We therefore do not claim that a frozen encoder makes a classifier more robust.

\subsection{Stability, determinism and cost}
\label{sec:cost}
\paragraph{Stability.} Over both learning rates, \nCollFt{} of \nCollFtN{} FT runs, \nCollLora{} of \nCollLoraN{} LoRA runs and \nCollPft{} of 264 PFT runs collapsed. After the learning-rate choice on dev, one of the 132 reported LoRA seeds of the main study (512 to 8,192 labels; the low-label extension is reported in \S\ref{sec:lowlabel}) stays collapsed and no FT or PFT seed does. \wsp{} has no optimisation trajectory: none of its \nCollWspN{} runs over all recipe versions collapsed.
\paragraph{Determinism.} For a fixed program, data subset, hardware and software stack, re-running \wsp{} reproduced every prediction file exactly (73 of 73 files across 23 re-run configurations). The editor is outside this claim: a new editor round can produce a different program.
\paragraph{Cost.} \wsp{} reads 17 encoder passes per input at inference where the trained models read one, and the headline results use all 17. Adapting is not faster. Timed on an idle GPU (Appendix~\ref{app:compute}), one v8 run on SST-2 took 8.9 minutes with 512 labels and 19.0 minutes with 8,192, against 1.97 minutes for one FT run (step-matched, so about the same at every budget): 2.3 and 4.8 times the two-run learning-rate search of a tuned baseline. Storage depends on the selected head. Per-view LDA heads take a few megabytes per task and the PCA-based heads about 0.2 GB, but kernel heads must keep the training features of all 216 views: about 0.45 GB per task with 512 labels and 7.2 GB with 8,192 (float32), against 1.4 GB for a fine-tuned copy of RoBERTa-large and about 3 MB for a LoRA adapter. At the larger budgets the selected program is usually a kernel head. A new task also needs what fine-tuning needs (the training labels and the dev set), and new templates need an editor round.

\begin{table}[t]
\centering\small
\setlength{\tabcolsep}{4pt}
\begin{tabularx}{\columnwidth}{l>{\raggedright\arraybackslash}Xrrr}
\toprule
& & \multicolumn{3}{c}{Labels} \\
\cmidrule(lr){3-5}
Version & What it adds & 512 & 2,048 & 8,192 \\
\midrule
v2 & label-word prior; template-pooled heads & 93.07 & 93.76 & 94.60 \\
v3 & prior-mean kernel; Occam's window & 93.20 & 93.78 & 94.73 \\
v4 & additive multi-view kernel & 93.21 & 93.82 & 94.73 \\
v5 & cross-fitted selection & 93.26 & 93.88 & 94.75 \\
v6 & new prompts for MPQA, TREC-6, SUBJ & 93.50 & 93.97 & 94.81 \\
v7 & new prompts for CR & 93.55 & 94.00 & 94.88 \\
\rowcolor{wspblue}v8 & 16-template prior; 4 more templates as views & 93.77 & 94.29 & 94.99 \\
\bottomrule
\end{tabularx}
\caption{The recipe ladder: average test accuracy (\%) of every WSP version on the development datasets (11, 10 and 11 datasets). Each version was written down with a time stamp before its own results and keeps everything above it; every version was scored on these test sets, and the steps have no intervals. The shaded row is the final recipe. v6 and v7 rewrote prompts using dev data only, for datasets chosen with their test results in view. The first recipe, R, exists for one data subset of a few datasets only (Appendix~\ref{app:prereg}).}
\label{tab:ladder}
\end{table}

\subsection{What mattered, and what did not}
\label{sec:ladder}
Table~\ref{tab:ladder} shows the recipe ladder: each version was fixed, with its time stamp, before its own results, and every version was scored on the development test sets, which is why these are development results. The steps were built in one order and their gains have no intervals, so they show what each step added in this sequence, not a decomposition. The steps that bought accuracy changed the program around the encoder. \emph{The label-word prior} (v2) entered every later recipe. \emph{Cross-fitted selection} (v5) adds little in domain but raises the out-of-domain accuracy of the SST-2 programs at 8,192 labels from 82.1\% to 92.5\% on IMDB and Yelp (a gap seen on the out-of-domain test sets): the dev set alone could not tell a robust head from a brittle one of equal dev accuracy. \emph{Dev-only editor rounds} (v6, v7) rewrote the prompts of four datasets ($+$1.35 on MPQA at 512 labels), but the tasks to revise were chosen with their test results in view. \emph{More templates} (v8: a label-word prior over 16 templates, four of which are added as views by train-only cross-validation) gave $+$0.22, $+$0.29 and $+$0.11 points at 512, 2,048 and 8,192 labels, but nothing on the held-out datasets ($-$0.13 and $+$0.05). Further templates gave nothing more: a further batch in new framings and a batch that names the classes in the prompt, each added only where train-only cross-validation favoured it ($-$0.003 and $-$0.03 on fresh data subsets).

After v7, every new idea went through a registered screen on dev and cross-fitted train rows, never the test set (from the demonstrations round on, on a fresh data subset), and only one passed: v8. Demonstrations in the input, the one idea that feeds labelled examples through the frozen network, came close three times (on v7, on v8, and on v8 restricted to inputs where every demonstration fits uncut: $+$0.17, $+$0.06 and $+$0.06 against bars of $+$0.20, $+$0.10 and $+$0.10), with small gains on short inputs and none on long ones. Among the 27 ideas that did not help (Appendix~\ref{app:negative}, which marks the ideas judged on test data) are closed-form edits of the encoder's feed-forward (FFN) weights, attention-head amplification, closed-form self-training on unlabelled text, demonstrations in the input, tangent-kernel views of the kind that describe prompt-based fine-tuning in the few-shot regime \citep{malladi2023kernel}, views from inside the network, all sixteen templates as views, class-offset calibration and greedy ensemble selection \citep{caruana2004ensemble}. At 512 labels, the best head of each family we tried lands on about the same selection estimate (dev and cross-fitted train rows): for the readouts we tried, reading the frozen encoder differently did not move this plateau, while more labels and better prompts did.

\section{Related work}
\label{sec:related}
\paragraph{Fine-tuning and its alternatives.} Full fine-tuning \citep{devlin2019bert,liu2019roberta} and parameter-efficient methods such as adapters \citep{houlsby2019parameter}, BitFit \citep{benzaken2022bitfit}, prompt tuning \citep{lester2021power} and LoRA \citep{hu2022lora} train per task; \wsp{} computes no gradient and trains no encoder weight. Fine-tuning large encoders on small data is unstable \citep{dodge2020fine,mosbach2021stability,zhang2021revisiting}, and it can distort pretrained features, which linear probing followed by fine-tuning avoids \citep{kumar2022finetuning}. In-context learning \citep{brown2020language} adapts without any fit at all, at the cost of a large model per prediction.

\paragraph{Readouts of frozen encoders.} Whether to fine-tune or to read frozen layers was studied early for contextual encoders: feature-based BERT with a good layer combination comes close to fine-tuning on some tasks \citep{devlin2019bert,peters2019tune}, and probe suites read frozen sentence representations \citep{conneau2018senteval}. Closer to us, prompt-augmented linear probing trains a linear head on prompted representations of a frozen language model \citep{cho2023prompt}, and PromptBERT reads the \mask{} state of a template as the representation \citep{jiang2022promptbert}; our LP-prompt baseline is a single-template version of the former. kNN prompting classifies by nearest neighbours over prompted outputs \citep{xu2023knn}. Closed-form ridge and discriminant heads on frozen backbones are used in few-shot and continual learning \citep{bertinetto2019metalearning,mcdonnell2023ranpac}. What \wsp{} adds to this line is a selected program over many prompted views, layers, head families and a label-word prior, cross-fitted on the labelled data and compared with tuned fine-tuning under one protocol.

\paragraph{Gradient-free prompted classifiers.} Black-box tuning searches continuous prompts of a frozen model without gradients \citep{sun2022bbt,sun2022bbtv2}, and PromptBoosting boosts prompt-based weak classifiers built from a frozen model's outputs with few forward passes \citep{hou2023promptboosting}; both report results close to fine-tuning in few-shot settings. \wsp{} differs in reading hidden states as well as label-word scores and in fitting its heads in closed form rather than searching.

\paragraph{Prompts, verbalizers and calibration.} Cloze templates turn classification into masked-word prediction \citep{schick2021exploiting,gao2021making}; templates and label words can be searched automatically \citep{shin2020autoprompt,schick2020automatically} or expanded with knowledge \citep{hu2022knowledgeable}, and ensembles of prompts \citep{jiang2020know,schick2021exploiting} and calibration \citep{zhao2021calibrate} make zero-shot scores more reliable. Many of our templates and label words follow LM-BFF and PET \citep{gao2021making,schick2021exploiting} (Appendix~\ref{app:templates}). We use prompts as views of a frozen encoder and as a zero-shot prior; PFT, which fine-tunes through them, is our strongest baseline.

\paragraph{Training-free adaptation in vision.} Tip-Adapter mixes the zero-shot classifier of a frozen CLIP model with a cache kernel over few-shot training features \citep{zhang2022tip}, and Gaussian discriminant analysis on frozen CLIP features is a hard-to-beat training-free baseline \citep{wang2024hard}; LLM-written class descriptions improve such zero-shot classifiers \citep{menon2023visual,pratt2023platypus}. \wsp{}'s mixture and prior-mean kernel heads are close relatives of these designs, carried over to masked language models with prompted views.

\paragraph{LLMs that design programs.} LLMs have been used to write and optimise prompts \citep{zhou2023large,yang2024large} and to search over programs \citep{romeraparedes2024mathematical}, a form of automated machine learning \citep{hutter2019automated}. Our editor is used in that spirit, but we do not evaluate it as a search method.

\paragraph{Editing and programming weights.} Closed-form edits of factual associations change a network's weights \citep{meng2022locating,meng2023mass}. In weight-space programming as introduced, an LLM reads and writes a network's weights directly \citep{takehi2026sparks}. Concurrently with us, \citet{ok2026weightspace} let LLMs write the weights of small models by hand, with no fitting on data: on the combined ARC-Easy and ARC-Challenge test sets of the AI2 Reasoning Challenge their best written model reaches 54.4\%, against 82.1\% for a fine-tuned pretrained model of 0.8B parameters (Qwen3.5-0.8B), but its weights are readable and editable. Our variant writes no weight; it writes a program that computes a classifier's weights in closed form from labelled data on a frozen pretrained encoder. On our development datasets this comes within 0.5 points of the fine-tuning baselines, with no gradient step; on new tasks it stays behind prompt-based fine-tuning. We also tried editing the encoder's own weights in closed form (Appendix~\ref{app:negative}); for classification, fitting the classifier's weights on read activations mattered far more.

\section{Conclusion}
A closed-form program over prompted views of a frozen encoder, written with an LLM as design assistant, is better than tuned full fine-tuning and LoRA in our RoBERTa-large study: above both on 11 development datasets at every label budget, by more with 64 and 128 labels and most with 16 labels per class, and ahead of LoRA and within about half a point of FT on six held-out datasets, with no gradient step, no collapse and exact reproducibility for a fixed program. It is competitive with single-template prompt-based fine-tuning on the development datasets, within half a point, which shaped the recipe; on the held-out datasets, the cleaner but small check, PFT is ahead, by more than half a point with 8,192 labels, and every trained baseline is ahead on CoLA. What moved it were how heads are selected and how the task is phrased in the prompt; on new tasks, the remaining gap to prompt-based fine-tuning was not closed by any frozen-side idea we tried.

\section*{Limitations}
\begin{itemize}
\item \textbf{Development results.} The comparisons on the 11 development datasets informed the recipe (Appendix~\ref{app:prereg}); they are development results, not confirmatory ones.
\item \textbf{PFT is not matched, and not beaten.} PFT reads one template where \wsp{} reads eight and a 16-template prior; a multi-template PFT was not run. On the held-out datasets PFT is ahead (by 0.51 points with 512 labels and 1.30 with 8,192, the latter beyond the margin), most of all on CoLA and, with 8,192 labels, on the 5- and 10-class datasets.
\item \textbf{Held-out check.} Six datasets give little power, their task mix differs from the development datasets, and they were run in two batches (\S\ref{sec:heldout}).
\item \textbf{Acceptability and short opinion phrases.} \wsp{} trails every trained baseline on CoLA and is about one point behind PFT on MPQA and CR with 512 labels.
\item \textbf{Cost.} 17 encoder passes per input, slower adaptation than a tuned baseline, and gigabytes of stored features for kernel heads at 8,192 labels (\S\ref{sec:cost}).
\item \textbf{Scope.} One main encoder, RoBERTa-large. The final recipe matched FT and LoRA on roberta-base (within 0.12 points) but was 0.46 to 1.72 points behind them on bert-large (Appendix~\ref{app:encoders}); English single-text tasks with one-token label words only; the few-shot extension has one budget (16 labels per class); small test sets (CR: 376, TREC-6: 500, SST-2: 872) make differences under one point noisy.
\item \textbf{The editor.} The LLM-written programs were not compared with human-written or published templates, and an editor round is not deterministic.
\end{itemize}

\section*{Ethics statement}
We use public English text-classification datasets under their research licences and release no new data. \wsp{} needs no gradient training, but its inference cost (17 encoder passes per input) falls on whoever deploys it, and its kernel heads store training features, which matters when the training texts are sensitive. The LLM editor (Claude, Anthropic, through the Claude Code agent) wrote the programs, templates and code, ran the experiments and drafted analyses and text (Appendix~\ref{app:editor}); it read no test data while writing a program, and the authors checked and are responsible for everything reported.

\section*{Availability}
Code, the recipe programs and templates, the indices of every data subset, the time-stamped plan with every registration and deviation, and the prediction files will be released.

\bibliography{custom}

\appendix
\section{Datasets and setup}
\label{app:setup}
\begin{table}[htbp]
\centering\small\setlength{\tabcolsep}{3pt}
\rowcolors{2}{white}{rowgray}
\begin{tabular}{lrrrr}
\toprule
\hiderowcolors
Dataset & Classes & Train pool & Test & Max.\ tokens \\
\midrule
\showrowcolors
SST-2 & 2 & 66k & 872 & 64 \\
AG News & 4 & 119k & 7,600 & 128 \\
MR & 2 & 7.5k & 1,066 & 64 \\
SUBJ & 2 & 7.0k & 2,000 & 64 \\
CR & 2 & 2.4k & 376 & 96 \\
MPQA & 2 & 7.8k & 2,000 & 32 \\
TREC-6 & 6 & 4.4k & 500 & 48 \\
BBC News & 5 & 981 & 1,000 & 256 \\
Yelp-2 & 2 & 49k & 5,000$^*$ & 256 \\
Amazon-2 & 2 & 49k & 5,000$^*$ & 192 \\
IMDB & 2 & 24k & 5,000$^*$ & 256 \\
\midrule
SST-5 & 5 & 8.6k$^\ddagger$ & 2,210 & 64 \\
CoLA & 2 & 7.5k & 1,043 & 64 \\
DBpedia-14 & 14 & 50k$^\dagger$ & 5,000$^*$ & 128 \\
Yahoo-10 & 10 & 50k$^\dagger$ & 5,000$^*$ & 256 \\
Yelp-5 & 5 & 50k$^\dagger$ & 5,000$^*$ & 256 \\
Amazon-5 & 5 & 50k$^\dagger$ & 5,000$^*$ & 192 \\
\bottomrule
\end{tabular}
\caption{Datasets. Train pool: what remains after the 1,024-example dev split (BBC News: 245). $^*$A fixed class-stratified sample of the official test set. $^\dagger$A random sample of the training set. $^\ddagger$The official train and dev sets combined, before the dev split. CoLA's test set is the GLUE validation set (69.1\% acceptable).}
\label{tab:datasets}
\end{table}

\paragraph{Budgets and subsets.} Every dataset is run with 512 and 8,192 labelled examples, and all development datasets but BBC News also with 2,048. When the pool is smaller than the budget (MR, SUBJ, CR, MPQA, TREC-6, BBC News and CoLA at 8,192) the whole pool is used, and the cell has one data subset. Subset $k$ is drawn with seed $k$; FT, LoRA and PFT seed $k$ train on exactly subset $k$. Every method also uses the dev split (1,024 examples, BBC News 245) for selection, so the labelled data per cell is the budget plus the dev split.

\paragraph{Held-out datasets.} SST-5 and CoLA complete the single-sentence tasks of the LM-BFF suite \citep{gao2021making}; DBpedia-14, Yahoo-10, Yelp-5 and Amazon-5 complete the suite of \citet{zhang2015character}. Their round-1 programs were written blind, and a mechanical rule, fixed beforehand, gave a task one dev-only editor round if its zero-shot dev accuracy was below 85\%; five of the six met it (DBpedia-14 did not). Their programs are blind to their test data, but not to the literature: several of their templates and label words are those of LM-BFF or close to PET patterns (Appendix~\ref{app:templates}). The second batch (Yahoo-10, Yelp-5, Amazon-5) was run after the first batch's results were known; it had been named as optional from the start.

\paragraph{Trained baselines.} FT trains every parameter with a linear classification head; LoRA uses rank 8, $\alpha=16$ and dropout 0.1 on the query and value projections. All use AdamW (weight decay 0.01, none on biases and layer norms), batch size 32, 6\% linear warm-up, gradient clipping at 1.0, and the same maximum length as \wsp{}. Training is step-matched: 768 optimizer steps, the number of steps of 3 epochs on 8,192 labels (48 epochs at 512 labels, 12 at 2,048). Where the whole pool is used because it is smaller than the budget, the epochs are set to give a similar count: 654 to 834 steps (3 epochs for MR, SUBJ, MPQA, CoLA and SST-5, 6 for TREC-6, 10 for CR, 25 for BBC News). Learning rates: FT and PFT $\{10^{-5}, 3\cdot10^{-5}\}$, LoRA $\{3\cdot10^{-4}, 10^{-3}\}$; dev accuracy is evaluated every 50 steps (100 at 8,192 labels), and the learning rate and checkpoint with the best dev accuracy are kept per seed. On the development datasets the selected learning rates were: FT $10^{-5}$ in 76 of 96 seeds; LoRA $3\cdot10^{-4}$ in 55 of 96; PFT $10^{-5}$ in 70 of 96. FT's frequent choice of the lower edge of its grid suggests that a smaller learning rate might have helped it; we did not extend the grid. The long-text datasets are trained with bf16 autocast. PFT wraps the input in the first template of the task's \wsp{} program, scores a class by the mean logit of its label words at \mask{} from the pretrained MLM head, and adds no parameter. For CR, whose program changed in v7, PFT is run with both prompts and the better one on dev is kept per seed. The development effort was unequal: \wsp{} went through eight recipe versions, editor rounds and 27 screened ideas, the baselines through one fixed protocol.

\paragraph{Probes.} LP is $\ell_2$-regularised logistic regression on the last layer of the plain input (the \texttt{<s>} state or the mean of the token states, chosen on dev); LP-prompt is the same on the \mask{} state of the first template, as prompt-augmented linear probing \citep{cho2023prompt} with one template. The regularisation strength (9 values) is chosen on dev. LP-ml and LP-prompt-ml, registered in the revision before any of their runs, read the top 12 layers of the plain input (24 views) or the \mask{} state of the 16 prior templates at the top 12 layers (192 views), each view reduced to principal components fitted on the training subset (256 or 32 per view, 6,144 features), with the same fitting and selection.

\section{The closed-form heads}
\label{app:heads}
\begin{table}[htbp]
\centering\small\setlength{\tabcolsep}{3pt}
\begin{tabularx}{\columnwidth}{l>{\raggedright\arraybackslash}Xr}
\toprule
Family & Head and grid & \# \\
\midrule
mv & per-view shrinkage LDA, average of the per-view log-posteriors; shrinkage $\{0.1, 0.3, 0.6\}$ & 3 \\
cat & LDA on the concatenation of per-view PCA-256 projections; shrinkage $\{0.1, 0.3, 0.6, 0.8, 0.9, 0.95\}$, alone or averaged with mv$_{0.1}$ or mv$_{0.3}$ & 18 \\
stack & LDA on 5-fold cross-fitted per-view LDA log-posteriors; shrinkage $\{0.1, 0.3, 0.6\}$ & 3 \\
tpool & one LDA per (layer, pooling) fitted on the samples of all templates; shrinkage $\{0.1, 0.3, 0.6\}$ & 3 \\
rff & random-Fourier-feature ridge ($D = 8{,}192$); bandwidth $g \in \{0.5, 1, 2\}$ $\times$ ridge $\{10^{-3}, 10^{-2}, 10^{-1}\}$ & 9 \\
klin & exact linear kernel ridge in dual form on the standardised concatenation of all views; ridge $\{10^{-3}, 10^{-2}, 10^{-1}, 1\}$ & 4 \\
krbf & exact RBF kernel ridge, same features; $g \times$ ridge as rff & 9 \\
kadd & additive multi-view RBF kernel $\frac{1}{V}\sum_v \exp(-\lVert x_v - x'_v\rVert^2 / g\,m_v)$; $g \times$ ridge as rff & 9 \\
kprior, kaddprior & krbf or kadd with the prior as its mean: fit $Y-\beta P_0$, predict $\beta P_0(x) + K(x)A$; $g \times$ ridge $\times$ $\beta \in \{0.5, 1\}$ & 36 \\
zs & calibrated zero-shot label-word scores of the MLM head, summed over the 16 templates & 1 \\
zsfit & LDA on the label-word scores of all templates; shrinkage $\{0.1, 0.3\}$ & 2 \\
mix & each of the four supervised heads with the best dev accuracy, plus $\beta \times$ the log-posterior of zs or zsfit$_{0.3}$, $\beta \in \{0.5, 1\}$ & 16 \\
\midrule
& total & 113 \\
\bottomrule
\end{tabularx}
\caption{Candidate heads and their grids (113 per run). Bandwidths are relative to the median distance $m_v$; ridge values are relative to the trace of the kernel. Calibration of zs: the label-word scores minus their mean over the training subset \citep{zhao2021calibrate}.}
\label{tab:heads}
\end{table}

\paragraph{Selection.} The held-out rows of the selection are the dev rows and the 5-fold cross-fitted rows of the labelled training set. The stacked head's cross-fitted score is optimistic (its meta-head is fitted on posteriors whose own fits saw the other folds; 0.67 points above a fully nested cross-fit on average in control runs). This can only change which head is selected, never a reported test accuracy, since selection never reads test data.

\section{Results for every dataset}
\label{app:tables}
Tables~\ref{tab:app-dev-512}--\ref{tab:app-held-8192} give every method on every dataset with standard deviations, Table~\ref{tab:app-ni} the registered tests in full with the complementary statistics, Table~\ref{tab:app-probes} the multi-view probes of \S\ref{sec:probes} on every dataset, and Table~\ref{tab:cola} CoLA's Matthews correlation. Per-run comparisons: in the 33 paired runs per baseline at 512 labels on the development datasets (\wsp{} on subset $k$ against the baseline seed trained on subset $k$), \wsp{} is significantly better than FT in 9 and worse in 0 (McNemar's exact test, $p<0.05$), against LoRA 11 and 0, and against PFT 5 and 6; with 8,192 labels the counts are 3 and 0, 4 and 0, and 0 and 2.
\begin{table*}[tp]
\centering\small
\setlength{\tabcolsep}{4pt}
\rowcolors{1}{white}{rowgray}
\begin{tabular}{lrrrrrrrr}
\toprule
\hiderowcolors
& \multicolumn{5}{c}{\textit{frozen encoder, no gradient training}} & \multicolumn{3}{c}{\textit{gradient training}} \\
\cmidrule(lr){2-6}\cmidrule(lr){7-9}
Dataset & Zero-shot & LP & LP-prompt & WSP v7 & \textbf{WSP v8} & FT & LoRA & PFT \\
\midrule
\showrowcolors
SST-2 & 92.74 & 82.87\,{\color{black}\scriptsize$\pm$1.47} & 89.79\,{\color{black}\scriptsize$\pm$0.81} & 93.04\,{\color{black}\scriptsize$\pm$0.35} & 93.65\,{\color{black}\scriptsize$\pm$0.19} & 92.85\,{\color{black}\scriptsize$\pm$0.70} & 92.97\,{\color{black}\scriptsize$\pm$0.87} & 94.00\,{\color{black}\scriptsize$\pm$0.29} \\
AG News & 87.91 & 88.11\,{\color{black}\scriptsize$\pm$0.36} & 88.38\,{\color{black}\scriptsize$\pm$0.58} & 89.46\,{\color{black}\scriptsize$\pm$0.48} & 90.10\,{\color{black}\scriptsize$\pm$0.41} & 90.11\,{\color{black}\scriptsize$\pm$0.28} & 89.63\,{\color{black}\scriptsize$\pm$0.37} & 89.92\,{\color{black}\scriptsize$\pm$0.66} \\
MR & 90.46 & 80.68\,{\color{black}\scriptsize$\pm$0.58} & 86.34\,{\color{black}\scriptsize$\pm$0.04} & 90.15\,{\color{black}\scriptsize$\pm$0.47} & 90.68\,{\color{black}\scriptsize$\pm$0.85} & 88.81\,{\color{black}\scriptsize$\pm$0.51} & 88.49\,{\color{black}\scriptsize$\pm$0.38} & 90.81\,{\color{black}\scriptsize$\pm$0.15} \\
SUBJ & 81.58 & 94.08\,{\color{black}\scriptsize$\pm$0.52} & 95.35\,{\color{black}\scriptsize$\pm$0.47} & 96.28\,{\color{black}\scriptsize$\pm$0.02} & 96.38\,{\color{black}\scriptsize$\pm$0.16} & 95.50\,{\color{black}\scriptsize$\pm$0.43} & 96.20\,{\color{black}\scriptsize$\pm$0.14} & 96.62\,{\color{black}\scriptsize$\pm$0.28} \\
CR & 87.94 & 84.22\,{\color{black}\scriptsize$\pm$0.88} & 90.78\,{\color{black}\scriptsize$\pm$0.25} & 92.91\,{\color{black}\scriptsize$\pm$0.50} & 92.73\,{\color{black}\scriptsize$\pm$0.82} & 92.55\,{\color{black}\scriptsize$\pm$0.57} & 92.82\,{\color{black}\scriptsize$\pm$0.57} & 93.62\,{\color{black}\scriptsize$\pm$0.43} \\
MPQA & 83.23 & 75.78\,{\color{black}\scriptsize$\pm$0.52} & 81.33\,{\color{black}\scriptsize$\pm$0.84} & 87.23\,{\color{black}\scriptsize$\pm$0.50} & 87.37\,{\color{black}\scriptsize$\pm$0.36} & 87.40\,{\color{black}\scriptsize$\pm$0.25} & 88.28\,{\color{black}\scriptsize$\pm$0.46} & 88.68\,{\color{black}\scriptsize$\pm$0.25} \\
TREC-6 & 63.47 & 85.27\,{\color{black}\scriptsize$\pm$1.37} & 93.67\,{\color{black}\scriptsize$\pm$1.52} & 96.47\,{\color{black}\scriptsize$\pm$0.34} & 96.20\,{\color{black}\scriptsize$\pm$0.82} & 95.27\,{\color{black}\scriptsize$\pm$0.50} & 96.60\,{\color{black}\scriptsize$\pm$0.57} & 95.47\,{\color{black}\scriptsize$\pm$1.15} \\
BBC News & 93.60 & 96.53\,{\color{black}\scriptsize$\pm$0.61} & 96.37\,{\color{black}\scriptsize$\pm$0.25} & 97.43\,{\color{black}\scriptsize$\pm$0.33} & 97.47\,{\color{black}\scriptsize$\pm$0.25} & 97.50\,{\color{black}\scriptsize$\pm$0.29} & 97.47\,{\color{black}\scriptsize$\pm$0.61} & 97.90\,{\color{black}\scriptsize$\pm$0.75} \\
Yelp-2 & 94.85 & 93.19\,{\color{black}\scriptsize$\pm$0.19} & 91.58\,{\color{black}\scriptsize$\pm$0.21} & 96.70\,{\color{black}\scriptsize$\pm$0.03} & 96.85\,{\color{black}\scriptsize$\pm$0.09} & 96.71\,{\color{black}\scriptsize$\pm$0.36} & 96.73\,{\color{black}\scriptsize$\pm$0.18} & 96.68\,{\color{black}\scriptsize$\pm$0.21} \\
Amazon-2 & 94.64 & 90.55\,{\color{black}\scriptsize$\pm$0.70} & 90.87\,{\color{black}\scriptsize$\pm$0.56} & 95.65\,{\color{black}\scriptsize$\pm$0.05} & 96.15\,{\color{black}\scriptsize$\pm$0.14} & 95.51\,{\color{black}\scriptsize$\pm$0.12} & 95.43\,{\color{black}\scriptsize$\pm$0.12} & 95.49\,{\color{black}\scriptsize$\pm$0.35} \\
IMDB & 91.43 & 88.07\,{\color{black}\scriptsize$\pm$0.43} & 86.53\,{\color{black}\scriptsize$\pm$0.50} & 93.72\,{\color{black}\scriptsize$\pm$0.11} & 93.91\,{\color{black}\scriptsize$\pm$0.14} & 93.18\,{\color{black}\scriptsize$\pm$0.15} & 92.97\,{\color{black}\scriptsize$\pm$0.25} & 93.18\,{\color{black}\scriptsize$\pm$0.44} \\
\midrule
\rowcolor{white}
\textit{Average} (11) & 87.44 & 87.21\,{\scriptsize\phantom{$\pm$0.00}} & 90.09\,{\scriptsize\phantom{$\pm$0.00}} & 93.55\,{\scriptsize\phantom{$\pm$0.00}} & 93.77\,{\scriptsize\phantom{$\pm$0.00}} & 93.22\,{\scriptsize\phantom{$\pm$0.00}} & 93.42\,{\scriptsize\phantom{$\pm$0.00}} & 93.85\,{\scriptsize\phantom{$\pm$0.00}} \\
\bottomrule
\end{tabular}
\caption{Development datasets, 512 labels: test accuracy (\%). Small numbers: standard deviation (divisor 3) over 3 data subsets (WSP) or 3 seeds on the same subsets (baselines; collapsed seeds left out). A whole-pool cell has one subset, and WSP's has no deviation. Zero-shot: calibrated label-word scores (calibration on the training subset); LP and LP-prompt: last-layer probes.}
\label{tab:app-dev-512}
\end{table*}

\begin{table*}[tp]
\centering\small
\setlength{\tabcolsep}{4pt}
\rowcolors{1}{white}{rowgray}
\begin{tabular}{lrrrrrrrr}
\toprule
\hiderowcolors
& \multicolumn{5}{c}{\textit{frozen encoder, no gradient training}} & \multicolumn{3}{c}{\textit{gradient training}} \\
\cmidrule(lr){2-6}\cmidrule(lr){7-9}
Dataset & Zero-shot & LP & LP-prompt & WSP v7 & \textbf{WSP v8} & FT & LoRA & PFT \\
\midrule
\showrowcolors
SST-2 & 92.89 & 86.35\,{\color{black}\scriptsize$\pm$0.58} & 91.48\,{\color{black}\scriptsize$\pm$0.66} & 94.53\,{\color{black}\scriptsize$\pm$0.29} & 94.72\,{\color{black}\scriptsize$\pm$0.09} & 93.88\,{\color{black}\scriptsize$\pm$0.33} & 94.19\,{\color{black}\scriptsize$\pm$0.24} & 94.53\,{\color{black}\scriptsize$\pm$0.84} \\
AG News & 87.90 & 90.14\,{\color{black}\scriptsize$\pm$0.22} & 90.35\,{\color{black}\scriptsize$\pm$0.55} & 91.46\,{\color{black}\scriptsize$\pm$0.23} & 91.60\,{\color{black}\scriptsize$\pm$0.34} & 91.76\,{\color{black}\scriptsize$\pm$0.19} & 91.11\,{\color{black}\scriptsize$\pm$0.15} & 91.28\,{\color{black}\scriptsize$\pm$0.56} \\
MR & 90.56 & 84.93\,{\color{black}\scriptsize$\pm$0.50} & 87.87\,{\color{black}\scriptsize$\pm$0.44} & 91.12\,{\color{black}\scriptsize$\pm$0.42} & 91.46\,{\color{black}\scriptsize$\pm$0.20} & 90.59\,{\color{black}\scriptsize$\pm$0.16} & 90.28\,{\color{black}\scriptsize$\pm$0.71} & 91.78\,{\color{black}\scriptsize$\pm$0.62} \\
SUBJ & 81.65 & 95.73\,{\color{black}\scriptsize$\pm$0.38} & 95.97\,{\color{black}\scriptsize$\pm$0.17} & 97.17\,{\color{black}\scriptsize$\pm$0.12} & 97.30\,{\color{black}\scriptsize$\pm$0.04} & 97.05\,{\color{black}\scriptsize$\pm$0.04} & 96.88\,{\color{black}\scriptsize$\pm$0.16} & 97.30\,{\color{black}\scriptsize$\pm$0.25} \\
CR & 88.03 & 87.41\,{\color{black}\scriptsize$\pm$0.66} & 92.55\,{\color{black}\scriptsize$\pm$0.43} & 93.35\,{\color{black}\scriptsize$\pm$0.22} & 93.79\,{\color{black}\scriptsize$\pm$0.13} & 93.71\,{\color{black}\scriptsize$\pm$0.13} & 93.71\,{\color{black}\scriptsize$\pm$0.25} & 94.77\,{\color{black}\scriptsize$\pm$0.33} \\
MPQA & 83.17 & 80.92\,{\color{black}\scriptsize$\pm$1.33} & 83.68\,{\color{black}\scriptsize$\pm$0.44} & 88.17\,{\color{black}\scriptsize$\pm$0.17} & 88.57\,{\color{black}\scriptsize$\pm$0.41} & 89.07\,{\color{black}\scriptsize$\pm$0.53} & 89.00\,{\color{black}\scriptsize$\pm$0.52} & 89.20\,{\color{black}\scriptsize$\pm$0.90} \\
TREC-6 & 64.47 & 91.53\,{\color{black}\scriptsize$\pm$0.84} & 96.40\,{\color{black}\scriptsize$\pm$0.49} & 97.40\,{\color{black}\scriptsize$\pm$0.43} & 97.73\,{\color{black}\scriptsize$\pm$0.19} & 97.20\,{\color{black}\scriptsize$\pm$0.16} & 97.20\,{\color{black}\scriptsize$\pm$0.28} & 97.20\,{\color{black}\scriptsize$\pm$0.28} \\
Yelp-2 & 94.94 & 94.51\,{\color{black}\scriptsize$\pm$0.19} & 92.86\,{\color{black}\scriptsize$\pm$0.07} & 97.05\,{\color{black}\scriptsize$\pm$0.11} & 97.26\,{\color{black}\scriptsize$\pm$0.06} & 96.92\,{\color{black}\scriptsize$\pm$0.28} & 96.80\,{\color{black}\scriptsize$\pm$0.32} & 97.14\,{\color{black}\scriptsize$\pm$0.03} \\
Amazon-2 & 94.62 & 92.36\,{\color{black}\scriptsize$\pm$0.34} & 92.61\,{\color{black}\scriptsize$\pm$0.10} & 95.83\,{\color{black}\scriptsize$\pm$0.12} & 96.26\,{\color{black}\scriptsize$\pm$0.11} & 95.73\,{\color{black}\scriptsize$\pm$0.43} & 95.97\,{\color{black}\scriptsize$\pm$0.06} & 95.76\,{\color{black}\scriptsize$\pm$0.27} \\
IMDB & 91.44 & 89.93\,{\color{black}\scriptsize$\pm$0.40} & 87.35\,{\color{black}\scriptsize$\pm$0.31} & 93.91\,{\color{black}\scriptsize$\pm$0.02} & 94.15\,{\color{black}\scriptsize$\pm$0.14} & 93.58\,{\color{black}\scriptsize$\pm$0.19} & 93.55\,{\color{black}\scriptsize$\pm$0.17} & 93.92\,{\color{black}\scriptsize$\pm$0.16} \\
\midrule
\rowcolor{white}
\textit{Average} (10) & 86.97 & 89.38\,{\scriptsize\phantom{$\pm$0.00}} & 91.11\,{\scriptsize\phantom{$\pm$0.00}} & 94.00\,{\scriptsize\phantom{$\pm$0.00}} & 94.29\,{\scriptsize\phantom{$\pm$0.00}} & 93.95\,{\scriptsize\phantom{$\pm$0.00}} & 93.87\,{\scriptsize\phantom{$\pm$0.00}} & 94.29\,{\scriptsize\phantom{$\pm$0.00}} \\
\bottomrule
\end{tabular}
\caption{Development datasets, 2,048 labels: test accuracy (\%). Small numbers: standard deviation (divisor 3) over 3 data subsets (WSP) or 3 seeds on the same subsets (baselines; collapsed seeds left out). A whole-pool cell has one subset, and WSP's has no deviation. Zero-shot: calibrated label-word scores (calibration on the training subset); LP and LP-prompt: last-layer probes.}
\label{tab:app-dev-2048}
\end{table*}

\begin{table*}[tp]
\centering\small
\setlength{\tabcolsep}{4pt}
\rowcolors{1}{white}{rowgray}
\begin{tabular}{lrrrrrrrr}
\toprule
\hiderowcolors
& \multicolumn{5}{c}{\textit{frozen encoder, no gradient training}} & \multicolumn{3}{c}{\textit{gradient training}} \\
\cmidrule(lr){2-6}\cmidrule(lr){7-9}
Dataset & Zero-shot & LP & LP-prompt & WSP v7 & \textbf{WSP v8} & FT & LoRA & PFT \\
\midrule
\showrowcolors
SST-2 & 93.00 & 87.88\,{\color{black}\scriptsize$\pm$0.14} & 92.05\,{\color{black}\scriptsize$\pm$0.62} & 95.26\,{\color{black}\scriptsize$\pm$0.30} & 95.11\,{\color{black}\scriptsize$\pm$0.22} & 94.15\,{\color{black}\scriptsize$\pm$0.34} & 94.30\,{\color{black}\scriptsize$\pm$0.92} & 95.18\,{\color{black}\scriptsize$\pm$0.25} \\
AG News & 87.89 & 91.14\,{\color{black}\scriptsize$\pm$0.23} & 91.06\,{\color{black}\scriptsize$\pm$0.21} & 93.09\,{\color{black}\scriptsize$\pm$0.12} & 93.11\,{\color{black}\scriptsize$\pm$0.16} & 92.93\,{\color{black}\scriptsize$\pm$0.12} & 92.36\,{\color{black}\scriptsize$\pm$0.42} & 93.23\,{\color{black}\scriptsize$\pm$0.16} \\
MR & 90.43 & 86.30\,{\scriptsize\phantom{$\pm$0.00}} & 89.40\,{\scriptsize\phantom{$\pm$0.00}} & 91.28\,{\scriptsize\phantom{$\pm$0.00}} & 91.18\,{\scriptsize\phantom{$\pm$0.00}} & 91.34\,{\color{black}\scriptsize$\pm$0.19} & 91.34\,{\color{black}\scriptsize$\pm$0.44} & 91.84\,{\color{black}\scriptsize$\pm$1.03} \\
SUBJ & 81.70 & 96.80\,{\scriptsize\phantom{$\pm$0.00}} & 96.25\,{\scriptsize\phantom{$\pm$0.00}} & 97.40\,{\scriptsize\phantom{$\pm$0.00}} & 97.65\,{\scriptsize\phantom{$\pm$0.00}} & 97.38\,{\color{black}\scriptsize$\pm$0.08} & 97.27\,{\color{black}\scriptsize$\pm$0.18} & 97.65\,{\color{black}\scriptsize$\pm$0.08} \\
CR & 88.03 & 87.50\,{\scriptsize\phantom{$\pm$0.00}} & 92.82\,{\scriptsize\phantom{$\pm$0.00}} & 93.88\,{\scriptsize\phantom{$\pm$0.00}} & 94.15\,{\scriptsize\phantom{$\pm$0.00}} & 93.35\,{\color{black}\scriptsize$\pm$0.78} & 92.82\,{\color{black}\scriptsize$\pm$1.00} & 94.24\,{\color{black}\scriptsize$\pm$0.45} \\
MPQA & 83.10 & 84.20\,{\scriptsize\phantom{$\pm$0.00}} & 86.05\,{\scriptsize\phantom{$\pm$0.00}} & 89.90\,{\scriptsize\phantom{$\pm$0.00}} & 90.40\,{\scriptsize\phantom{$\pm$0.00}} & 89.90\,{\color{black}\scriptsize$\pm$0.57} & 89.70\,{\scriptsize\phantom{$\pm$0.00}} & 90.52\,{\color{black}\scriptsize$\pm$0.22} \\
TREC-6 & 64.40 & 93.40\,{\scriptsize\phantom{$\pm$0.00}} & 96.60\,{\scriptsize\phantom{$\pm$0.00}} & 97.20\,{\scriptsize\phantom{$\pm$0.00}} & 97.20\,{\scriptsize\phantom{$\pm$0.00}} & 96.60\,{\color{black}\scriptsize$\pm$0.49} & 97.20\,{\color{black}\scriptsize$\pm$0.16} & 97.00\,{\color{black}\scriptsize$\pm$0.16} \\
BBC News & 93.60 & 97.40\,{\scriptsize\phantom{$\pm$0.00}} & 96.60\,{\scriptsize\phantom{$\pm$0.00}} & 97.90\,{\scriptsize\phantom{$\pm$0.00}} & 97.90\,{\scriptsize\phantom{$\pm$0.00}} & 97.67\,{\color{black}\scriptsize$\pm$0.05} & 98.03\,{\color{black}\scriptsize$\pm$0.05} & 97.93\,{\color{black}\scriptsize$\pm$0.24} \\
Yelp-2 & 94.97 & 95.38\,{\color{black}\scriptsize$\pm$0.11} & 94.01\,{\color{black}\scriptsize$\pm$0.17} & 97.35\,{\color{black}\scriptsize$\pm$0.13} & 97.36\,{\color{black}\scriptsize$\pm$0.09} & 97.29\,{\color{black}\scriptsize$\pm$0.26} & 97.34\,{\color{black}\scriptsize$\pm$0.19} & 97.55\,{\color{black}\scriptsize$\pm$0.10} \\
Amazon-2 & 94.62 & 93.43\,{\color{black}\scriptsize$\pm$0.04} & 93.76\,{\color{black}\scriptsize$\pm$0.07} & 96.04\,{\color{black}\scriptsize$\pm$0.10} & 96.37\,{\color{black}\scriptsize$\pm$0.02} & 96.29\,{\color{black}\scriptsize$\pm$0.37} & 96.23\,{\color{black}\scriptsize$\pm$0.09} & 96.15\,{\color{black}\scriptsize$\pm$0.20} \\
IMDB & 91.45 & 91.08\,{\color{black}\scriptsize$\pm$0.10} & 89.35\,{\color{black}\scriptsize$\pm$0.19} & 94.37\,{\color{black}\scriptsize$\pm$0.08} & 94.49\,{\color{black}\scriptsize$\pm$0.08} & 94.27\,{\color{black}\scriptsize$\pm$0.15} & 94.35\,{\color{black}\scriptsize$\pm$0.13} & 94.57\,{\color{black}\scriptsize$\pm$0.12} \\
\midrule
\rowcolor{white}
\textit{Average} (11) & 87.56 & 91.32\,{\scriptsize\phantom{$\pm$0.00}} & 92.54\,{\scriptsize\phantom{$\pm$0.00}} & 94.88\,{\scriptsize\phantom{$\pm$0.00}} & 94.99\,{\scriptsize\phantom{$\pm$0.00}} & 94.65\,{\scriptsize\phantom{$\pm$0.00}} & 94.63\,{\scriptsize\phantom{$\pm$0.00}} & 95.08\,{\scriptsize\phantom{$\pm$0.00}} \\
\bottomrule
\end{tabular}
\caption{Development datasets, 8,192 labels (whole pool when smaller): test accuracy (\%). Small numbers: standard deviation (divisor 3) over 3 data subsets (WSP) or 3 seeds on the same subsets (baselines; collapsed seeds left out). A whole-pool cell has one subset, and WSP's has no deviation (LoRA on MPQA: the mean of its 2 seeds that did not collapse, no deviation shown). Zero-shot: calibrated label-word scores (calibration on the training subset); LP and LP-prompt: last-layer probes.}
\label{tab:app-dev-8192}
\end{table*}

\begin{table*}[tp]
\centering\small
\setlength{\tabcolsep}{4pt}
\rowcolors{1}{white}{rowgray}
\begin{tabular}{lrrrrrrrr}
\toprule
\hiderowcolors
& \multicolumn{5}{c}{\textit{frozen encoder, no gradient training}} & \multicolumn{3}{c}{\textit{gradient training}} \\
\cmidrule(lr){2-6}\cmidrule(lr){7-9}
Dataset & Zero-shot & LP & LP-prompt & WSP v7 & \textbf{WSP v8} & FT & LoRA & PFT \\
\midrule
\showrowcolors
SST-5 & 44.16 & 41.75\,{\color{black}\scriptsize$\pm$1.82} & 51.19\,{\color{black}\scriptsize$\pm$0.85} & 56.30\,{\color{black}\scriptsize$\pm$0.64} & 56.20\,{\color{black}\scriptsize$\pm$1.09} & 53.14\,{\color{black}\scriptsize$\pm$1.83} & 52.31\,{\color{black}\scriptsize$\pm$0.52} & 55.44\,{\color{black}\scriptsize$\pm$0.92} \\
CoLA & 45.13 & 78.01\,{\color{black}\scriptsize$\pm$0.12} & 72.96\,{\color{black}\scriptsize$\pm$0.28} & 78.78\,{\color{black}\scriptsize$\pm$0.45} & 78.62\,{\color{black}\scriptsize$\pm$0.55} & 80.31\,{\color{black}\scriptsize$\pm$0.80} & 80.12\,{\color{black}\scriptsize$\pm$0.65} & 80.09\,{\color{black}\scriptsize$\pm$0.16} \\
DBpedia-14 & 93.31 & 95.30\,{\color{black}\scriptsize$\pm$0.28} & 98.17\,{\color{black}\scriptsize$\pm$0.02} & 98.33\,{\color{black}\scriptsize$\pm$0.15} & 98.31\,{\color{black}\scriptsize$\pm$0.23} & 98.39\,{\color{black}\scriptsize$\pm$0.12} & 98.61\,{\color{black}\scriptsize$\pm$0.16} & 98.59\,{\color{black}\scriptsize$\pm$0.18} \\
Yahoo-10 & 64.50 & 64.40\,{\color{black}\scriptsize$\pm$1.12} & 70.15\,{\color{black}\scriptsize$\pm$0.42} & 69.37\,{\color{black}\scriptsize$\pm$0.36} & 69.49\,{\color{black}\scriptsize$\pm$0.67} & 69.18\,{\color{black}\scriptsize$\pm$0.48} & 69.31\,{\color{black}\scriptsize$\pm$0.39} & 70.60\,{\color{black}\scriptsize$\pm$0.57} \\
Yelp-5 & 55.84 & 51.54\,{\color{black}\scriptsize$\pm$0.23} & 61.13\,{\color{black}\scriptsize$\pm$0.41} & 62.45\,{\color{black}\scriptsize$\pm$0.29} & 62.27\,{\color{black}\scriptsize$\pm$0.54} & 61.53\,{\color{black}\scriptsize$\pm$0.63} & 60.24\,{\color{black}\scriptsize$\pm$1.22} & 63.63\,{\color{black}\scriptsize$\pm$0.51} \\
Amazon-5 & 51.81 & 47.00\,{\color{black}\scriptsize$\pm$0.29} & 59.40\,{\color{black}\scriptsize$\pm$0.49} & 59.99\,{\color{black}\scriptsize$\pm$0.33} & 59.57\,{\color{black}\scriptsize$\pm$0.80} & 58.71\,{\color{black}\scriptsize$\pm$0.27} & 57.07\,{\color{black}\scriptsize$\pm$0.69} & 59.19\,{\color{black}\scriptsize$\pm$1.18} \\
\midrule
\rowcolor{white}
\textit{Average} (6) & 59.12 & 63.00\,{\scriptsize\phantom{$\pm$0.00}} & 68.83\,{\scriptsize\phantom{$\pm$0.00}} & 70.87\,{\scriptsize\phantom{$\pm$0.00}} & 70.74\,{\scriptsize\phantom{$\pm$0.00}} & 70.21\,{\scriptsize\phantom{$\pm$0.00}} & 69.61\,{\scriptsize\phantom{$\pm$0.00}} & 71.26\,{\scriptsize\phantom{$\pm$0.00}} \\
\bottomrule
\end{tabular}
\caption{Held-out datasets, 512 labels: test accuracy (\%), as in Table~\ref{tab:app-dev-512}. The held-out datasets were added after the recipe ladder up to v7 was fixed; their programs were fixed before any run on them, and each recipe version (v2 to v7, later v8) was run on them once.}
\label{tab:app-held-512}
\end{table*}

\begin{table*}[tp]
\centering\small
\setlength{\tabcolsep}{4pt}
\rowcolors{1}{white}{rowgray}
\begin{tabular}{lrrrrrrrr}
\toprule
\hiderowcolors
& \multicolumn{5}{c}{\textit{frozen encoder, no gradient training}} & \multicolumn{3}{c}{\textit{gradient training}} \\
\cmidrule(lr){2-6}\cmidrule(lr){7-9}
Dataset & Zero-shot & LP & LP-prompt & WSP v7 & \textbf{WSP v8} & FT & LoRA & PFT \\
\midrule
\showrowcolors
SST-5 & 44.18 & 50.36\,{\color{black}\scriptsize$\pm$0.10} & 57.32\,{\color{black}\scriptsize$\pm$0.55} & 59.41\,{\color{black}\scriptsize$\pm$0.20} & 59.80\,{\color{black}\scriptsize$\pm$0.37} & 58.84\,{\color{black}\scriptsize$\pm$0.95} & 58.05\,{\color{black}\scriptsize$\pm$0.69} & 60.87\,{\color{black}\scriptsize$\pm$0.64} \\
CoLA & 45.25 & 81.30\,{\scriptsize\phantom{$\pm$0.00}} & 77.76\,{\scriptsize\phantom{$\pm$0.00}} & 83.03\,{\scriptsize\phantom{$\pm$0.00}} & 82.74\,{\scriptsize\phantom{$\pm$0.00}} & 85.08\,{\color{black}\scriptsize$\pm$0.36} & 83.73\,{\color{black}\scriptsize$\pm$0.82} & 85.08\,{\color{black}\scriptsize$\pm$0.55} \\
DBpedia-14 & 93.32 & 98.29\,{\color{black}\scriptsize$\pm$0.05} & 98.95\,{\color{black}\scriptsize$\pm$0.08} & 99.24\,{\color{black}\scriptsize$\pm$0.03} & 99.28\,{\color{black}\scriptsize$\pm$0.03} & 99.09\,{\color{black}\scriptsize$\pm$0.09} & 99.03\,{\color{black}\scriptsize$\pm$0.12} & 99.19\,{\color{black}\scriptsize$\pm$0.10} \\
Yahoo-10 & 64.46 & 71.49\,{\color{black}\scriptsize$\pm$0.15} & 73.06\,{\color{black}\scriptsize$\pm$0.24} & 74.07\,{\color{black}\scriptsize$\pm$0.19} & 73.76\,{\color{black}\scriptsize$\pm$0.45} & 74.61\,{\color{black}\scriptsize$\pm$0.31} & 74.32\,{\color{black}\scriptsize$\pm$0.13} & 75.17\,{\color{black}\scriptsize$\pm$0.28} \\
Yelp-5 & 55.97 & 59.55\,{\color{black}\scriptsize$\pm$0.50} & 63.05\,{\color{black}\scriptsize$\pm$0.62} & 64.93\,{\color{black}\scriptsize$\pm$0.06} & 65.33\,{\color{black}\scriptsize$\pm$0.18} & 65.79\,{\color{black}\scriptsize$\pm$0.76} & 65.13\,{\color{black}\scriptsize$\pm$0.71} & 66.99\,{\color{black}\scriptsize$\pm$0.49} \\
Amazon-5 & 52.07 & 56.12\,{\color{black}\scriptsize$\pm$0.09} & 61.96\,{\color{black}\scriptsize$\pm$0.09} & 62.37\,{\color{black}\scriptsize$\pm$0.05} & 62.45\,{\color{black}\scriptsize$\pm$0.12} & 62.30\,{\color{black}\scriptsize$\pm$0.44} & 61.84\,{\color{black}\scriptsize$\pm$1.28} & 63.87\,{\color{black}\scriptsize$\pm$0.33} \\
\midrule
\rowcolor{white}
\textit{Average} (6) & 59.21 & 69.52\,{\scriptsize\phantom{$\pm$0.00}} & 72.02\,{\scriptsize\phantom{$\pm$0.00}} & 73.84\,{\scriptsize\phantom{$\pm$0.00}} & 73.89\,{\scriptsize\phantom{$\pm$0.00}} & 74.28\,{\scriptsize\phantom{$\pm$0.00}} & 73.68\,{\scriptsize\phantom{$\pm$0.00}} & 75.19\,{\scriptsize\phantom{$\pm$0.00}} \\
\bottomrule
\end{tabular}
\caption{Held-out datasets, 8,192 labels: test accuracy (\%), as in Table~\ref{tab:app-dev-512}. The held-out datasets were added after the recipe ladder up to v7 was fixed; their programs were fixed before any run on them, and each recipe version (v2 to v7, later v8) was run on them once.}
\label{tab:app-held-8192}
\end{table*}

\begin{table*}[tp]
\centering\small
\setlength{\tabcolsep}{4pt}
\rowcolors{1}{white}{rowgray}
\begin{tabular}{lllrrrcrcc}
\toprule
\hiderowcolors
Datasets & Labels & vs & $k$ & Mean diff. & SD & 90\% interval & Lower bound & Non-inferior ($p$) & Inferior ($p$) \\
\midrule
\showrowcolors
Development & 512 & FT & 11 & $+$0.56 & 0.58 & [$+$0.24, $+$0.87] & $+$0.29 & yes ($<$0.001) & no (1.000) \\
 &  & LoRA & 11 & $+$0.35 & 0.81 & [$-$0.09, $+$0.80] & $-$0.02 & yes (0.003) & no (0.997) \\
 &  & PFT & 11 & $-$0.08 & 0.66 & [$-$0.44, $+$0.28] & $-$0.40 & yes (0.031) & no (0.969) \\
Development & 2,048 & FT & 10 & $+$0.34 & 0.43 & [$+$0.08, $+$0.59] & $+$0.12 & yes ($<$0.001) & no (1.000) \\
 &  & LoRA & 10 & $+$0.42 & 0.41 & [$+$0.18, $+$0.65] & $+$0.21 & yes ($<$0.001) & no (1.000) \\
 &  & PFT & 10 & $\pm$0.00 & 0.49 & [$-$0.29, $+$0.28] & $-$0.26 & yes (0.006) & no (0.994) \\
Development & 8,192 & FT & 11 & $+$0.34 & 0.34 & [$+$0.16, $+$0.53] & $+$0.19 & yes ($<$0.001) & no (1.000) \\
 &  & LoRA & 11 & $+$0.36 & 0.48 & [$+$0.10, $+$0.62] & $+$0.14 & yes ($<$0.001) & no (1.000) \\
 &  & PFT & 11 & $-$0.09 & 0.23 & [$-$0.21, $+$0.04] & $-$0.20 & yes ($<$0.001) & no (1.000) \\
Held-out & 512 & FT & 6 & $+$0.53 & 1.54 & [$-$0.74, $+$1.80] & $-$0.40 & no (0.081) & no (0.919) \\
 &  & LoRA & 6 & $+$1.13 & 2.01 & [$-$0.52, $+$2.79] & $-$0.08 & no (0.052) & no (0.949) \\
 &  & PFT & 6 & $-$0.51 & 0.94 & [$-$1.29, $+$0.26] & $-$1.08 & no (0.515) & no (0.487) \\
Held-out & 8,192 & FT & 6 & $-$0.39 & 1.14 & [$-$1.33, $+$0.54] & $-$1.11 & no (0.412) & no (0.588) \\
 &  & LoRA & 6 & $+$0.21 & 0.96 & [$-$0.58, $+$1.00] & $-$0.37 & no (0.064) & no (0.936) \\
 &  & PFT & 6 & $-$1.30 & 0.80 & [$-$1.96, $-$0.64] & $-$1.76 & no (0.971) & yes (0.029) \\
\bottomrule
\end{tabular}
\caption{The registered non-inferiority tests in full, with the complementary statistics added in the revision. $k$: datasets; Mean diff.: mean over datasets of WSP minus the baseline (points); SD: sample standard deviation of the per-dataset differences; 90\% interval: two-sided $t$-interval. Lower bound: 5th percentile of 10,000 bootstrap resamples of datasets. Non-inferior: one-sided paired $t$-test of $H_0$: mean difference $\le -0.5$ with $p < 0.05$, and the lower bound above $-0.5$ (the registered test). Inferior: the complementary one-sided test of $H_0$: mean difference $\ge -0.5$. No correction for the 15 comparisons; with Bonferroni ($\alpha = 0.05/15$) the development tests against PFT at 512 and 2,048 labels would not pass. With six held-out datasets, the registered test can pass for a truly level method only if the SD is below about 0.61.}
\label{tab:app-ni}
\end{table*}

\begin{table*}[tp]
\centering\small
\setlength{\tabcolsep}{3pt}
\rowcolors{1}{white}{rowgray}
\begin{tabular}{lrrrrrrrrrrrr}
\toprule
\hiderowcolors
& \multicolumn{4}{c}{\textbf{512 labels}} & \multicolumn{4}{c}{\textbf{2,048 labels}} & \multicolumn{4}{c}{\textbf{8,192 labels}} \\
\cmidrule(lr){2-5}\cmidrule(lr){6-9}\cmidrule(lr){10-13}
Dataset & LP-p & LP-ml & LP-p-ml & \textbf{WSP} & LP-p & LP-ml & LP-p-ml & \textbf{WSP} & LP-p & LP-ml & LP-p-ml & \textbf{WSP} \\
\midrule
\showrowcolors
SST-2 & 89.79 & 87.16 & 92.81 & \textbf{93.65} & 91.48 & 90.29 & 93.96 & \textbf{94.72} & 92.05 & 91.70 & 94.00 & \textbf{95.11} \\
AG News & 88.38 & 88.32 & 88.10 & \textbf{90.10} & 90.35 & 90.25 & 89.82 & \textbf{91.60} & 91.06 & 91.44 & 91.43 & \textbf{93.11} \\
MR & 86.34 & 83.52 & 89.99 & \textbf{90.68} & 87.87 & 87.71 & 91.03 & \textbf{91.46} & 89.40 & 88.93 & 90.90 & \textbf{91.18} \\
SUBJ & 95.35 & 95.35 & 95.67 & \textbf{96.38} & 95.97 & 96.92 & 96.88 & \textbf{97.30} & 96.25 & 97.05 & 96.95 & \textbf{97.65} \\
CR & 90.78 & 88.30 & 91.76 & \textbf{92.73} & 92.55 & 91.05 & 92.46 & \textbf{93.79} & 92.82 & 90.43 & 92.55 & \textbf{94.15} \\
MPQA & 81.33 & 80.75 & 84.75 & \textbf{87.37} & 83.68 & 84.92 & 86.87 & \textbf{88.57} & 86.05 & 86.90 & 88.85 & \textbf{90.40} \\
TREC-6 & 93.67 & 89.40 & 95.13 & \textbf{96.20} & 96.40 & 93.60 & 97.00 & \textbf{97.73} & 96.60 & 95.60 & 97.00 & \textbf{97.20} \\
BBC News & 96.37 & 96.90 & 95.17 & \textbf{97.47} & -- & -- & -- & -- & 96.60 & 97.20 & 96.00 & \textbf{97.90} \\
Yelp-2 & 91.58 & 94.93 & 95.89 & \textbf{96.85} & 92.86 & 95.77 & 96.33 & \textbf{97.26} & 94.01 & 96.41 & 97.06 & \textbf{97.36} \\
Amazon-2 & 90.87 & 93.19 & 95.25 & \textbf{96.15} & 92.61 & 94.40 & 95.51 & \textbf{96.26} & 93.76 & 95.09 & 96.12 & \textbf{96.37} \\
IMDB & 86.53 & 89.25 & 92.49 & \textbf{93.91} & 87.35 & 91.20 & 93.22 & \textbf{94.15} & 89.35 & 92.61 & 93.97 & \textbf{94.49} \\
\midrule
\rowcolor{white}
\textit{Average} & 90.09 & 89.73 & 92.46 & \textbf{93.77} & 91.11 & 91.61 & 93.31 & \textbf{94.29} & 92.54 & 93.03 & 94.07 & \textbf{94.99} \\
\midrule
\rowcolor{white}\multicolumn{13}{l}{\textit{held-out datasets}} \\
SST-5 & 51.19 & 42.08 & 52.31 & \textbf{56.20} & -- & -- & -- & -- & 57.32 & 51.79 & 58.17 & \textbf{59.80} \\
CoLA & 72.96 & 76.96 & 74.11 & \textbf{78.62} & -- & -- & -- & -- & 77.76 & 81.98 & 79.58 & \textbf{82.74} \\
DBpedia-14 & 98.17 & 96.63 & 97.65 & \textbf{98.31} & -- & -- & -- & -- & 98.95 & 98.73 & 98.70 & \textbf{99.28} \\
Yahoo-10 & \textbf{70.15} & 65.98 & 64.81 & 69.49 & -- & -- & -- & -- & 73.06 & 71.79 & 71.62 & \textbf{73.76} \\
Yelp-5 & 61.13 & 50.85 & 59.32 & \textbf{62.27} & -- & -- & -- & -- & 63.05 & 60.29 & 63.89 & \textbf{65.33} \\
Amazon-5 & 59.40 & 48.43 & 56.02 & \textbf{59.57} & -- & -- & -- & -- & 61.96 & 57.03 & 61.36 & \textbf{62.45} \\
\midrule
\rowcolor{white}
\textit{Average} & 68.83 & 63.49 & 67.37 & \textbf{70.74} & -- & -- & -- & -- & 72.02 & 70.27 & 72.22 & \textbf{73.89} \\
\bottomrule
\end{tabular}
\caption{The stronger probes (registered and added in the revision, \S\ref{sec:probes}) against WSP, test accuracy (\%), means over 3 data subsets (one whole-pool run where the pool is smaller than 8,192). LP-p (LP-prompt): last-layer \mask{} state of PFT's template. LP-ml: the top 12 layers of the plain input, both poolings (24 views); LP-p-ml (LP-prompt-ml): the \mask{} state of the 16 templates of WSP's prior in the top 12 layers (192 views); each view reduced to principal components fit on the training subset (6,144 features), then the same logistic regression with the regularisation chosen on dev. \textbf{Bold}: the best of the four.}
\label{tab:app-probes}
\end{table*}

\begin{table}[htbp]
\centering\small\setlength{\tabcolsep}{4pt}
\begin{tabular}{lrrrr}
\toprule
Labels & \textbf{WSP} & FT & LoRA & PFT \\
\midrule
512 & 46.8 & 51.0 & 50.6 & 51.0 \\
8,192 (whole pool) & 58.2 & 63.8 & 60.2 & 63.8 \\
\bottomrule
\end{tabular}
\caption{CoLA: Matthews correlation ($\times 100$) on the GLUE validation set, computed from the saved test predictions (mean over 3 subsets or seeds; WSP at 8,192 labels: one whole-pool run). The majority rate is 69.1\%, and the zero-shot prior reaches 45.1\% accuracy.}
\label{tab:cola}
\end{table}

\section{Negative results}
\label{app:negative}
\begin{table*}[tp]
\centering\small
\rowcolors{2}{white}{rowgray}
\begin{tabularx}{\textwidth}{>{\raggedright\arraybackslash}p{5.2cm}X>{\raggedright\arraybackslash}p{2.6cm}}
\toprule
\hiderowcolors
Idea & Evidence & Outcome \\
\midrule
\showrowcolors
Attention-head amplification$^\dagger$ & test $+$0.3, mostly out of domain & not adopted \\
Closed-form FFN key$\to$value edits of weak neurons$^\dagger$ & dev $+$1.4, test $-$1.6 to $-$2.0 & negative \\
Multi-round weight edits accepted on held-out train rows$^\dagger$ & test $-$0.2 to $+$0.4 & negative \\
Closed-form self-training on 32k unlabelled texts & never selected on dev & negative \\
8 templates instead of 4 (recipe R) & no gain at 8,192 labels & negative \\
Multi-encoder kernels & dev gains did not transfer & negative \\
Auto-verbalizer (top-$k$ MLM words per class, chosen on train) & dev: at or below the manual label words & negative \\
Diagonal shrinkage target (z-scored per-view LDA) & dev $-$0.4 to $+$0.4, sign flips & neutral \\
IDF-weighted / max token pooling & dev $-$0.5 to $+$0.3 / $-$2.0 to $-$0.1 & negative \\
Reading lower layers (windows 1--12, 7--18, 1--24) & layers 13--24 best or within 1 point on 6 of 6 datasets & negative \\
Semi-supervised LDA (covariance from train + unlabelled dev inputs) & dev 85.8--88.1 vs.\ 89.0--89.9 & negative \\
Template editor round on AG News & dev $+$0.6, below the 1-point rule & negative \\
Editor round 3: MPQA again; AG News label words & dev $+$0.6; $+$0.9 and $+$0.6 & negative \\
Read depth per budget (3 / 6 / 12 layers) & held-out estimate highest with 12 layers & negative \\
Tangent views: kernel ridge on per-example gradients w.r.t.\ all bias vectors & below the best existing head on 11 of 11 datasets (mean $-$1.4) & negative \\
View selection (top 8 / 16 / 32 views by inner cross-validation) & mean $-$0.05, ahead on 4 of 11 & negative \\
Views from inside the network at \mask{} (FFN neurons, MLM-head state, vocabulary log-probabilities) & $+$0.04 over the standard views (rule: $+$0.2) & negative \\
Demonstrations in the input (as LM-BFF; fold-clean cross-fitting) & $+$0.17 (ahead on 7 of 11), dev-only $-$0.03 (rule: $+$0.2, dev-only $\ge 0$) & negative \\
Larger label-word prior (16 templates); with view choice among the 16 & $+$0.03; $+$0.11 (rule: $+$0.2) & negative alone; v8 combines the larger prior with added views \\
Tangent views with every weight (sketched per-example gradients: closed-form linearised PFT) & best head $-$1.20 vs.\ the best existing head; added: $-$0.16 & negative \\
Class-offset calibration of the selected program & honest half-split gain $-$0.13, positive on 0 of 11 & negative \\
All 16 templates as views (408 views) & $-$0.06 vs.\ v8 (ahead on 5 of 11), dev-only $-$0.14 & negative \\
Greedy ensemble selection \citep{caruana2004ensemble} & nested gain $+$0.08, positive on 6 of 11 (rule: $+$0.10, 7) & negative \\
A further template batch in new framings, added greedily where train-only CV favours it & $-$0.003, behind on 6 of 11 (rule: $+$0.10, behind $\le 3$) & negative \\
Templates that name the classes (``\textit{x} Was it terrible or great ? \mask{} .''), added the same way & $-$0.03, ahead on 5 of 11 (rule: $+$0.10) & negative \\
Demonstrations on top of v8 (as above) & $+$0.06, ahead on 7 of 11, dev-only $+$0.04 (rule: $+$0.10) & negative \\
Demonstrations only where every one fits uncut (6 short-input datasets) & $+$0.06 over 11 datasets; MR $+$0.35, MPQA $+$0.30, CR $-$0.12 (rule: $+$0.10) & negative \\
\bottomrule
\end{tabularx}
\caption{The 27 ideas that did not help, with their evidence, in the order they were tried. $^\dagger$Judged partly on test results (the first, weight-editing phase of the project); every later idea was judged on dev data or on the held-out estimate (dev and cross-fitted train rows). The registered rounds (from the demonstrations on) were scored before any test run, on dev and cross-fitted train rows of a fresh data subset at 512 labels, against a control arm. The ideas that were adopted are not listed here; they are the recipe versions of Table~\ref{tab:prereg}. The tangent-view results concern the closed-form kernels of our recipe at 512 labels, not the few-shot regime in which \citet{malladi2023kernel} describe prompt-based fine-tuning as kernel regression.}
\label{tab:negative}
\end{table*}
Table~\ref{tab:negative} lists every idea that did not help. A methodological lesson from the demonstrations round: when labelled training items enter the features (here as demonstrations in the input of other items), cross-fitted scores leak a held-out item's label through its neighbours' demonstrations. In our runs this raised cross-fitted accuracy by 0.4 to 0.6 points on average over the candidates, and by up to 6.8 points for single heads, until every cross-fitted score was rebuilt fold-clean.

\section{Other encoders}
\label{app:encoders}
We ran the recipe, unchanged, on roberta-base and bert-large (SST-2 and AG News, 8,192 labels, 3 subsets), with the same baseline protocol, first with v7 and then with v8; each check was registered before any of its runs. Table~\ref{tab:encoders} gives the v8 results. On roberta-base, \wsp{} is within 0.12 points of the better of FT and LoRA, as on roberta-large, and PFT is ahead (by 1.03 points on SST-2 and 0.37 on AG News). On bert-large the result does not transfer: \wsp{} is 1.72 points behind LoRA on SST-2 and 0.46 behind FT on AG News. On bert-large SST-2 we also ran the encoder-specific program of a registered editor round, which is v7 on this encoder; with v8's added templates and views it is 0.95 points below its v7 result (91.55) and 2.06 behind LoRA, so v8's additions did not help on this encoder. The recipe reads the encoder through cloze prompts and the \mask{} state, a signal much stronger in RoBERTa than in BERT.
\begin{table*}[tp]
\centering\small\setlength{\tabcolsep}{5pt}
\begin{tabular}{llcccccc}
\toprule
Encoder & Dataset & \wsp{} v8 & v7 & FT & LoRA & PFT & $\Delta_{\text{FT/LoRA}}$ \\
\midrule
roberta-base & SST-2 & 92.20\,{\scriptsize$\pm$0.16} & 91.93 & 91.82 & 92.16 & 93.23 & $+$0.04 \\
roberta-base & AG News & 92.60\,{\scriptsize$\pm$0.13} & 92.63 & 92.72 & 91.91 & 92.97 & $-$0.12 \\
bert-large & SST-2 & 90.94\,{\scriptsize$\pm$0.41} & 90.94 & 92.24 & 92.66 & 92.28 & $-$1.72 \\
bert-large & SST-2$^{e}$ & 90.60\,{\scriptsize$\pm$0.82} & 91.55 & 92.24 & 92.66 & 92.28 & $-$2.06 \\
bert-large & AG News & 92.07\,{\scriptsize$\pm$0.05} & 92.01 & 92.54 & 92.08 & 92.42 & $-$0.46 \\
\bottomrule
\end{tabular}
\caption{Other encoders, 8,192 labels, test accuracy (\%), mean (and standard deviation for \wsp{}) over 3 subsets. $\Delta_{\text{FT/LoRA}}$: \wsp{} v8 minus the better of FT and LoRA, from unrounded means. $^{e}$The registered encoder-specific editor-round program (v7's on bert-large SST-2); every other row is the unchanged recipe, for which v7 equals v5 on these two datasets.}
\label{tab:encoders}
\end{table*}

\section{Low-label extension: 64 and 128 labels}
\label{app:lowlabel}
\paragraph{Design.} The extension was registered on 22 November 2026, before any of its runs, as an evaluation of the final recipe, not a recipe change: \wsp{} is v8, and nothing in it was changed in response to these results. It covers the 11 development datasets with 64 and 128 labelled training examples, on 3 stratified data subsets (seed $k$ = subset $k$, as in the main study). FT, LoRA and PFT use the study's learning-rate grids and train for 768 optimizer steps at batch size 32 (384 epochs at 64 labels, 192 at 128), with the dev set evaluated every 50 steps and the learning rate and checkpoint with the best dev accuracy kept per seed; CR's PFT is run with both of its programs, and the better one on dev is kept per seed. LP and LP-prompt are run as in the main study. The analysis was fixed with the design: the per-dataset and average tables and the registered non-inferiority tests of \S\ref{sec:setup} at each budget against FT, LoRA and PFT, reported whatever they show.

\paragraph{Two disclosures.} (i) \emph{The dev set.} The dev split stays the study's fixed 1,024 examples (BBC News: 245), larger than the training set at these budgets. It is the same for every method and is used for selection only, but it means the methods see more labelled data than the budget names; a setting with a dev set as small as the training set is the separate few-shot extension (Appendix~\ref{app:fewshot}). (ii) \emph{A crash fix.} \wsp{} on TREC-6 at 64 labels, subset 0, stopped with an error: the stacked LDA head's linear system was singular. Its five cross-fitting folds are not stratified, and with 64 labels a TREC-6 class can have a single example, so the training part of a fold can miss a class, whose mean was then undefined. The fix gives a class absent from an LDA's training rows the overall mean as its class mean and floors the class priors at $10^{-6}$; when every class is present, the computation is unchanged bit for bit. We checked every data subset: the fix changes TREC-6 only, whose rarest class has one or two training examples at these budgets and is missing from some cross-fitting part in all six of its subsets. On the other ten datasets at 64 and 128 labels, and on all 17 datasets at 512 labels, the smallest budget of the main study, every class is present in every part. The failed run had produced no result; it was rerun with the fixed code, which every TREC-6 run used.

Tables~\ref{tab:app-lo-64} and~\ref{tab:app-lo-128} give every method with standard deviations (no baseline seed stayed collapsed after the learning-rate choice), and Table~\ref{tab:app-ni-lo} the registered tests in full.
\begin{table*}[tp]
\centering\small
\setlength{\tabcolsep}{4pt}
\rowcolors{3}{white}{rowgray}
\begin{tabular}{lrrrrrrr}
\toprule
& \multicolumn{4}{c}{\textit{frozen encoder, no gradient training}} & \multicolumn{3}{c}{\textit{gradient training}} \\
\cmidrule(lr){2-5}\cmidrule(lr){6-8}
Dataset & Zero-shot & LP & LP-prompt & \textbf{WSP v8} & FT & LoRA & PFT \\
\midrule
SST-2 & 92.89 & 69.92\,{\color{black}\scriptsize$\pm$6.38} & 86.54\,{\color{black}\scriptsize$\pm$1.54} & 93.08\,{\color{black}\scriptsize$\pm$0.05} & 86.09\,{\color{black}\scriptsize$\pm$0.87} & 88.53\,{\color{black}\scriptsize$\pm$1.14} & 92.43\,{\color{black}\scriptsize$\pm$0.66} \\
AG News & 88.03 & 84.14\,{\color{black}\scriptsize$\pm$1.12} & 83.15\,{\color{black}\scriptsize$\pm$0.61} & 88.66\,{\color{black}\scriptsize$\pm$0.38} & 88.23\,{\color{black}\scriptsize$\pm$0.85} & 88.33\,{\color{black}\scriptsize$\pm$1.12} & 89.05\,{\color{black}\scriptsize$\pm$0.75} \\
MR & 90.40 & 72.51\,{\color{black}\scriptsize$\pm$1.51} & 84.52\,{\color{black}\scriptsize$\pm$0.87} & 90.43\,{\color{black}\scriptsize$\pm$0.35} & 84.43\,{\color{black}\scriptsize$\pm$1.84} & 85.87\,{\color{black}\scriptsize$\pm$1.65} & 89.62\,{\color{black}\scriptsize$\pm$0.19} \\
SUBJ & 80.72 & 90.83\,{\color{black}\scriptsize$\pm$0.67} & 93.05\,{\color{black}\scriptsize$\pm$0.85} & 94.37\,{\color{black}\scriptsize$\pm$0.51} & 93.60\,{\color{black}\scriptsize$\pm$0.46} & 93.58\,{\color{black}\scriptsize$\pm$0.72} & 95.22\,{\color{black}\scriptsize$\pm$0.55} \\
CR & 87.15 & 70.48\,{\color{black}\scriptsize$\pm$1.36} & 88.21\,{\color{black}\scriptsize$\pm$0.45} & 90.16\,{\color{black}\scriptsize$\pm$1.42} & 82.80\,{\color{black}\scriptsize$\pm$4.02} & 87.06\,{\color{black}\scriptsize$\pm$5.13} & 92.82\,{\color{black}\scriptsize$\pm$1.00} \\
MPQA & 82.83 & 57.48\,{\color{black}\scriptsize$\pm$2.52} & 76.45\,{\color{black}\scriptsize$\pm$2.46} & 80.17\,{\color{black}\scriptsize$\pm$2.29} & 72.15\,{\color{black}\scriptsize$\pm$5.67} & 68.43\,{\color{black}\scriptsize$\pm$4.38} & 84.65\,{\color{black}\scriptsize$\pm$1.66} \\
TREC-6 & 63.60 & 61.33\,{\color{black}\scriptsize$\pm$7.36} & 86.00\,{\color{black}\scriptsize$\pm$5.40} & 90.20\,{\color{black}\scriptsize$\pm$3.83} & 81.53\,{\color{black}\scriptsize$\pm$12.12} & 83.93\,{\color{black}\scriptsize$\pm$10.58} & 90.53\,{\color{black}\scriptsize$\pm$2.58} \\
BBC News & 93.43 & 93.20\,{\color{black}\scriptsize$\pm$0.54} & 92.33\,{\color{black}\scriptsize$\pm$0.19} & 94.67\,{\color{black}\scriptsize$\pm$0.90} & 95.90\,{\color{black}\scriptsize$\pm$0.36} & 95.50\,{\color{black}\scriptsize$\pm$0.37} & 96.57\,{\color{black}\scriptsize$\pm$0.48} \\
Yelp-2 & 94.73 & 88.89\,{\color{black}\scriptsize$\pm$0.52} & 88.31\,{\color{black}\scriptsize$\pm$1.06} & 95.43\,{\color{black}\scriptsize$\pm$0.38} & 95.87\,{\color{black}\scriptsize$\pm$0.17} & 95.56\,{\color{black}\scriptsize$\pm$0.22} & 94.90\,{\color{black}\scriptsize$\pm$1.16} \\
Amazon-2 & 94.53 & 83.11\,{\color{black}\scriptsize$\pm$2.55} & 88.69\,{\color{black}\scriptsize$\pm$0.69} & 94.95\,{\color{black}\scriptsize$\pm$0.23} & 94.18\,{\color{black}\scriptsize$\pm$0.63} & 94.22\,{\color{black}\scriptsize$\pm$0.83} & 93.81\,{\color{black}\scriptsize$\pm$0.32} \\
IMDB & 91.41 & 78.61\,{\color{black}\scriptsize$\pm$0.72} & 84.06\,{\color{black}\scriptsize$\pm$1.17} & 92.17\,{\color{black}\scriptsize$\pm$0.35} & 91.26\,{\color{black}\scriptsize$\pm$0.65} & 90.88\,{\color{black}\scriptsize$\pm$0.48} & 91.36\,{\color{black}\scriptsize$\pm$0.47} \\
\midrule
\rowcolor{white}
\textit{Average} (11) & 87.25 & 77.32\,{\scriptsize\phantom{$\pm$0.00}} & 86.48\,{\scriptsize\phantom{$\pm$0.00}} & 91.30\,{\scriptsize\phantom{$\pm$0.00}} & 87.82\,{\scriptsize\phantom{$\pm$0.00}} & 88.35\,{\scriptsize\phantom{$\pm$0.00}} & 91.91\,{\scriptsize\phantom{$\pm$0.00}} \\
\bottomrule
\end{tabular}
\caption{Low-label extension, 64 labels: test accuracy (\%). Small numbers: standard deviation (divisor 3) over 3 data subsets (WSP, probes) or 3 seeds on the same subsets (FT, LoRA, PFT). No baseline seed stayed collapsed after the learning-rate choice on dev. Zero-shot: calibrated label-word scores (calibration on the training subset); LP and LP-prompt: last-layer probes.}
\label{tab:app-lo-64}
\end{table*}

\begin{table*}[tp]
\centering\small
\setlength{\tabcolsep}{4pt}
\rowcolors{3}{white}{rowgray}
\begin{tabular}{lrrrrrrr}
\toprule
& \multicolumn{4}{c}{\textit{frozen encoder, no gradient training}} & \multicolumn{3}{c}{\textit{gradient training}} \\
\cmidrule(lr){2-5}\cmidrule(lr){6-8}
Dataset & Zero-shot & LP & LP-prompt & \textbf{WSP v8} & FT & LoRA & PFT \\
\midrule
SST-2 & 93.00 & 74.58\,{\color{black}\scriptsize$\pm$3.56} & 87.96\,{\color{black}\scriptsize$\pm$0.82} & 92.70\,{\color{black}\scriptsize$\pm$0.92} & 89.95\,{\color{black}\scriptsize$\pm$1.52} & 91.82\,{\color{black}\scriptsize$\pm$0.52} & 92.93\,{\color{black}\scriptsize$\pm$0.97} \\
AG News & 88.04 & 86.43\,{\color{black}\scriptsize$\pm$0.45} & 86.61\,{\color{black}\scriptsize$\pm$0.26} & 88.89\,{\color{black}\scriptsize$\pm$0.58} & 89.21\,{\color{black}\scriptsize$\pm$0.27} & 89.06\,{\color{black}\scriptsize$\pm$0.09} & 89.43\,{\color{black}\scriptsize$\pm$0.40} \\
MR & 90.40 & 76.20\,{\color{black}\scriptsize$\pm$0.65} & 85.27\,{\color{black}\scriptsize$\pm$0.28} & 90.24\,{\color{black}\scriptsize$\pm$0.48} & 85.74\,{\color{black}\scriptsize$\pm$0.58} & 87.15\,{\color{black}\scriptsize$\pm$0.41} & 89.93\,{\color{black}\scriptsize$\pm$0.27} \\
SUBJ & 81.67 & 92.47\,{\color{black}\scriptsize$\pm$0.35} & 93.87\,{\color{black}\scriptsize$\pm$0.24} & 94.70\,{\color{black}\scriptsize$\pm$0.43} & 94.15\,{\color{black}\scriptsize$\pm$1.13} & 94.43\,{\color{black}\scriptsize$\pm$0.91} & 95.83\,{\color{black}\scriptsize$\pm$0.08} \\
CR & 87.68 & 76.60\,{\color{black}\scriptsize$\pm$0.75} & 89.54\,{\color{black}\scriptsize$\pm$0.55} & 91.93\,{\color{black}\scriptsize$\pm$0.50} & 87.94\,{\color{black}\scriptsize$\pm$1.95} & 90.87\,{\color{black}\scriptsize$\pm$1.09} & 93.17\,{\color{black}\scriptsize$\pm$0.25} \\
MPQA & 82.70 & 65.03\,{\color{black}\scriptsize$\pm$2.13} & 78.72\,{\color{black}\scriptsize$\pm$1.34} & 84.37\,{\color{black}\scriptsize$\pm$1.27} & 80.43\,{\color{black}\scriptsize$\pm$0.62} & 78.10\,{\color{black}\scriptsize$\pm$1.68} & 87.90\,{\color{black}\scriptsize$\pm$0.86} \\
TREC-6 & 63.67 & 72.13\,{\color{black}\scriptsize$\pm$3.64} & 88.00\,{\color{black}\scriptsize$\pm$4.38} & 92.80\,{\color{black}\scriptsize$\pm$1.07} & 91.20\,{\color{black}\scriptsize$\pm$2.83} & 92.07\,{\color{black}\scriptsize$\pm$1.05} & 93.93\,{\color{black}\scriptsize$\pm$1.93} \\
BBC News & 93.30 & 94.60\,{\color{black}\scriptsize$\pm$1.19} & 94.13\,{\color{black}\scriptsize$\pm$0.40} & 96.73\,{\color{black}\scriptsize$\pm$0.40} & 96.57\,{\color{black}\scriptsize$\pm$0.83} & 97.30\,{\color{black}\scriptsize$\pm$0.14} & 96.73\,{\color{black}\scriptsize$\pm$0.48} \\
Yelp-2 & 94.90 & 90.57\,{\color{black}\scriptsize$\pm$0.46} & 89.75\,{\color{black}\scriptsize$\pm$0.34} & 95.89\,{\color{black}\scriptsize$\pm$0.45} & 95.81\,{\color{black}\scriptsize$\pm$0.26} & 95.90\,{\color{black}\scriptsize$\pm$0.10} & 95.90\,{\color{black}\scriptsize$\pm$0.28} \\
Amazon-2 & 94.61 & 85.86\,{\color{black}\scriptsize$\pm$2.15} & 89.79\,{\color{black}\scriptsize$\pm$0.62} & 94.95\,{\color{black}\scriptsize$\pm$0.13} & 94.77\,{\color{black}\scriptsize$\pm$0.32} & 94.81\,{\color{black}\scriptsize$\pm$0.52} & 94.74\,{\color{black}\scriptsize$\pm$0.11} \\
IMDB & 91.42 & 83.61\,{\color{black}\scriptsize$\pm$0.97} & 85.46\,{\color{black}\scriptsize$\pm$0.19} & 92.90\,{\color{black}\scriptsize$\pm$0.23} & 91.84\,{\color{black}\scriptsize$\pm$0.62} & 91.65\,{\color{black}\scriptsize$\pm$0.21} & 90.71\,{\color{black}\scriptsize$\pm$0.74} \\
\midrule
\rowcolor{white}
\textit{Average} (11) & 87.40 & 81.64\,{\scriptsize\phantom{$\pm$0.00}} & 88.10\,{\scriptsize\phantom{$\pm$0.00}} & 92.37\,{\scriptsize\phantom{$\pm$0.00}} & 90.69\,{\scriptsize\phantom{$\pm$0.00}} & 91.20\,{\scriptsize\phantom{$\pm$0.00}} & 92.84\,{\scriptsize\phantom{$\pm$0.00}} \\
\bottomrule
\end{tabular}
\caption{Low-label extension, 128 labels: test accuracy (\%). Small numbers: standard deviation (divisor 3) over 3 data subsets (WSP, probes) or 3 seeds on the same subsets (FT, LoRA, PFT). No baseline seed stayed collapsed after the learning-rate choice on dev. Zero-shot: calibrated label-word scores (calibration on the training subset); LP and LP-prompt: last-layer probes.}
\label{tab:app-lo-128}
\end{table*}

\begin{table*}[tp]
\centering\small
\setlength{\tabcolsep}{4pt}
\rowcolors{2}{white}{rowgray}
\begin{tabular}{llrrrcrcc}
\toprule
Labels & vs & $k$ & Mean diff. & SD & 90\% interval & Lower bound & Non-inferior ($p$) & Inferior ($p$) \\
\midrule
64 & FT & 11 & $+$3.48 & 3.87 & [$+$1.36, $+$5.59] & $+$1.66 & yes (0.003) & no (0.997) \\
 & LoRA & 11 & $+$2.94 & 3.69 & [$+$0.93, $+$4.96] & $+$1.30 & yes (0.006) & no (0.994) \\
 & PFT & 11 & $-$0.61 & 1.76 & [$-$1.57, $+$0.36] & $-$1.48 & no (0.577) & no (0.423) \\
128 & FT & 11 & $+$1.68 & 1.80 & [$+$0.70, $+$2.66] & $+$0.84 & yes (0.001) & no (0.999) \\
 & LoRA & 11 & $+$1.18 & 1.95 & [$+$0.11, $+$2.24] & $+$0.35 & yes (0.009) & no (0.991) \\
 & PFT & 11 & $-$0.47 & 1.40 & [$-$1.23, $+$0.30] & $-$1.15 & no (0.468) & no (0.532) \\
\bottomrule
\end{tabular}
\caption{The registered tests of the low-label extension in full, as in Table~\ref{tab:app-ni}. Non-inferior: one-sided paired $t$-test of $H_0$: mean difference $\le -0.5$ (with the bootstrap bound, the registered test); Inferior: the complementary test of $H_0$: mean difference $\ge -0.5$. No multiplicity correction.}
\label{tab:app-ni-lo}
\end{table*}

\section{Few-shot extension: 16 labels per class}
\label{app:fewshot}
\paragraph{Design.} The extension was registered on 23 November 2026, before any of its runs, as an evaluation of the final recipe, not a recipe change: \wsp{} is v8 (with the fix of Appendix~\ref{app:lowlabel}), and nothing in it was changed in response to these results. Every one of the 17 datasets is run with $K=16$ labelled training examples per class and a dev set of 16 per class, drawn from the study's fixed dev split with its own seed, on 5 data subsets (seeds 0--4, as in LM-BFF); every method sees the same training and dev examples. \wsp{} selects on the small dev set plus its cross-fitted training rows, as always. FT, LoRA and PFT are trained as in the few-shot setting of LM-BFF: batch size 8, 1,000 optimizer steps, the dev set evaluated every 100 steps, and the learning rate (the study's grids) and checkpoint with the best dev accuracy kept per seed; CR's PFT is run with both of its programs, and the better on dev is kept per seed. The analysis was fixed with the design: per-dataset means and averages for the development and the held-out datasets separately, and the registered non-inferiority tests of \S\ref{sec:setup} with the complementary statistics.

\paragraph{Disclosures.} The runs were paused for about 13 hours to free the GPU for another project and resumed from the record of finished runs; the two runs in progress at the pause were rerun from the start. The analysis script was written after the runs, implementing the analysis fixed at registration; it breaks ties between learning rates in favour of the lower one. No baseline seed collapsed. Table~\ref{tab:app-fs} gives every dataset with standard deviations and the tests in full.
\begin{table*}[tp]
\centering\small
\setlength{\tabcolsep}{5pt}
\rowcolors{2}{white}{rowgray}
\begin{tabular}{lrrrr}
\toprule
Dataset & \textbf{WSP v8} & FT & LoRA & PFT \\
\midrule
SST-2 & 92.36\,{\scriptsize$\pm$0.65} & 77.20\,{\scriptsize$\pm$5.69} & 82.68\,{\scriptsize$\pm$3.56} & 89.93\,{\scriptsize$\pm$2.22} \\
AG News & 86.89\,{\scriptsize$\pm$0.95} & 87.42\,{\scriptsize$\pm$1.06} & 87.35\,{\scriptsize$\pm$0.77} & 88.62\,{\scriptsize$\pm$0.59} \\
MR & 87.97\,{\scriptsize$\pm$1.97} & 71.91\,{\scriptsize$\pm$6.45} & 74.41\,{\scriptsize$\pm$6.04} & 87.52\,{\scriptsize$\pm$3.02} \\
SUBJ & 91.56\,{\scriptsize$\pm$2.67} & 90.31\,{\scriptsize$\pm$2.52} & 91.05\,{\scriptsize$\pm$1.81} & 94.42\,{\scriptsize$\pm$0.50} \\
CR & 88.99\,{\scriptsize$\pm$1.77} & 79.36\,{\scriptsize$\pm$4.75} & 82.98\,{\scriptsize$\pm$4.80} & 92.55\,{\scriptsize$\pm$0.69} \\
MPQA & 81.21\,{\scriptsize$\pm$3.25} & 74.52\,{\scriptsize$\pm$2.69} & 73.97\,{\scriptsize$\pm$4.13} & 85.25\,{\scriptsize$\pm$2.31} \\
TREC-6 & 88.32\,{\scriptsize$\pm$2.11} & 82.52\,{\scriptsize$\pm$7.74} & 86.16\,{\scriptsize$\pm$4.70} & 84.52\,{\scriptsize$\pm$4.37} \\
BBC News & 94.98\,{\scriptsize$\pm$0.76} & 96.32\,{\scriptsize$\pm$0.80} & 96.48\,{\scriptsize$\pm$0.41} & 96.94\,{\scriptsize$\pm$0.48} \\
Yelp-2 & 95.18\,{\scriptsize$\pm$0.93} & 94.70\,{\scriptsize$\pm$0.53} & 93.98\,{\scriptsize$\pm$0.68} & 93.97\,{\scriptsize$\pm$1.44} \\
Amazon-2 & 93.68\,{\scriptsize$\pm$1.20} & 91.22\,{\scriptsize$\pm$2.79} & 91.67\,{\scriptsize$\pm$2.51} & 93.16\,{\scriptsize$\pm$0.86} \\
IMDB & 89.14\,{\scriptsize$\pm$1.90} & 86.36\,{\scriptsize$\pm$3.06} & 87.86\,{\scriptsize$\pm$3.49} & 88.56\,{\scriptsize$\pm$1.17} \\
\midrule
\rowcolor{white}
\textit{Average} (11) & 90.03\,{\scriptsize\phantom{$\pm$0.00}} & 84.71\,{\scriptsize\phantom{$\pm$0.00}} & 86.24\,{\scriptsize\phantom{$\pm$0.00}} & 90.50\,{\scriptsize\phantom{$\pm$0.00}} \\
\midrule
SST-5 & 47.18\,{\scriptsize$\pm$1.26} & 39.38\,{\scriptsize$\pm$2.31} & 40.66\,{\scriptsize$\pm$1.89} & 48.44\,{\scriptsize$\pm$3.50} \\
CoLA & 61.19\,{\scriptsize$\pm$3.36} & 69.88\,{\scriptsize$\pm$2.75} & 66.44\,{\scriptsize$\pm$2.43} & 57.78\,{\scriptsize$\pm$7.87} \\
DBpedia-14 & 97.42\,{\scriptsize$\pm$0.55} & 97.88\,{\scriptsize$\pm$0.32} & 98.28\,{\scriptsize$\pm$0.31} & 98.23\,{\scriptsize$\pm$0.28} \\
Yahoo-10 & 67.70\,{\scriptsize$\pm$0.74} & 65.96\,{\scriptsize$\pm$1.41} & 66.46\,{\scriptsize$\pm$1.10} & 69.69\,{\scriptsize$\pm$0.72} \\
Yelp-5 & 57.18\,{\scriptsize$\pm$2.63} & 50.94\,{\scriptsize$\pm$2.04} & 51.21\,{\scriptsize$\pm$3.76} & 59.16\,{\scriptsize$\pm$1.14} \\
Amazon-5 & 52.20\,{\scriptsize$\pm$2.45} & 44.60\,{\scriptsize$\pm$1.23} & 44.71\,{\scriptsize$\pm$2.35} & 57.87\,{\scriptsize$\pm$0.48} \\
\midrule
\rowcolor{white}
\textit{Average} (6) & 63.81\,{\scriptsize\phantom{$\pm$0.00}} & 61.44\,{\scriptsize\phantom{$\pm$0.00}} & 61.29\,{\scriptsize\phantom{$\pm$0.00}} & 65.20\,{\scriptsize\phantom{$\pm$0.00}} \\
\bottomrule
\end{tabular}
\par\medskip
\begin{tabular}{llrrrrrc}
\toprule
Datasets & Baseline & WSP $-$ baseline & 90\% interval & $p$ (non-inf.) & bootstrap 5\% & ahead / behind & passes \\
\midrule
development (11) & FT & $+$5.31 & [$+$2.01, $+$8.62] & 0.005 & $+$2.58 & 9 / 2 & \checkmark \\
 & LoRA & $+$3.79 & [$+$1.21, $+$6.37] & 0.006 & $+$1.67 & 9 / 2 & \checkmark \\
 & PFT & $-$0.47 & [$-$1.85, $+$0.92] & 0.484 & $-$1.69 & 6 / 5 & $\times$ \\
held-out (6) & FT & $+$2.37 & [$-$2.86, $+$7.60] & 0.160 & $-$1.75 & 4 / 2 & $\times$ \\
 & LoRA & $+$2.52 & [$-$1.61, $+$6.65] & 0.101 & $-$0.60 & 4 / 2 & $\times$ \\
 & PFT & $-$1.39 & [$-$3.78, $+$1.01] & 0.755 & $-$3.10 & 1 / 5 & $\times$ \\
\bottomrule
\end{tabular}
\caption{Few-shot extension (16 labels per class, dev 16 per class): test accuracy (\%) with the standard deviation (divisor 5) over 5 data subsets (WSP) or 5 seeds (baselines), and the registered tests across datasets as in Table~\ref{tab:ni}: non-inferiority at a 0.5-point margin (one-sided $t$-test and bootstrap lower bound; \checkmark: both clear the margin), the two-sided 90\% $t$-interval and the datasets ahead and behind. No baseline seed collapsed.}
\label{tab:app-fs}
\end{table*}

\section{Compute and storage}
\label{app:compute}
\paragraph{Time.} One RTX 5090 (32 GB), timed with the GPU otherwise idle (the other jobs on the machine paused) and checked against FT control runs. The v8 timing control was 21\% faster than its reference that day, so by a rule fixed before any v8 time was read, v8 is reported as a ratio to the FT control of the same queue, on the reference scale: on SST-2, one v8 run took 9.65 times one FT run at 8,192 labels (19.0 minutes on the reference scale; peak GPU memory 11.9 GB) and 4.51 times at 512 labels (8.9 minutes; 5.8 GB), where one FT run takes 1.97 minutes. A tuned baseline needs two runs (its learning-rate grid) per subset, so v8 takes 4.8 and 2.3 times as long. The AG News timings of the same queue are not used: the process that paused the other jobs stopped early, so those runs shared the GPU. Earlier, one v7 run (5 encoder passes) took 3 to 6.5 minutes and one training run 1.7 to 2.9 minutes. These times cover model loading, featurisation, all candidate heads and their selection, and the test evaluation; inference itself needs 17 encoder passes per input for v8 (5 for v7) where a trained model needs one.
\paragraph{Storage.} \wsp{} keeps one frozen encoder for all tasks plus a head per task. Per-view LDA heads take a few megabytes per task; the PCA-based heads keep 216 projection matrices (about 0.2 GB); the kernel heads keep the standardised training features of all 216 views, $n \times 221{,}184$ values: about 0.45 GB with 512 labels and 7.2 GB with 8,192 in float32. A fine-tuned copy of RoBERTa-large is 1.4 GB (355M parameters in float32) and a LoRA adapter about 3 MB (0.8M parameters).
\paragraph{Determinism.} Re-running \wsp{} with the same program, data subset, hardware and software reproduced every prediction file exactly (73 of 73 files across 23 re-run configurations).

\section{Protocol: what was fixed when, and every deviation}
\label{app:prereg}
\paragraph{The records.} Every recipe version, registered round and statistical test was written as a dated entry, with its lab-clock time, in a plan file kept during the project, before its own results were computed. The file was not under version control and was not deposited with a third-party registry, so its times cannot be verified independently; we release it unchanged with the code. We call this a time-stamped internal plan, not a pre-registration. Several time stamps were written wrongly, by minutes, and later corrected in the file with notes; none changes the order of events.
\begin{table}[htbp]
\centering\small\setlength{\tabcolsep}{3pt}
\begin{tabularx}{\columnwidth}{l>{\raggedright\arraybackslash}X}
\toprule
Recipe & Adds; adoption rule \\
\midrule
R & multi-view reading of 12 layers, kernel heads \\
v2 & MLM-head prior, template pooling, top-3 rule \\
v3 & prior-mean kernel, Occam's window \\
v4 & additive multi-view kernel \\
v5 & cross-fitted selection \\
v6 & editor round 2: dev-selected templates and label words for MPQA, TREC-6, SUBJ; rule: $\ge$1 point of \mask{}-LDA dev accuracy \\
v7 & editor round 3: CR; same rule \\
v8 & round 9: label-word prior over 16 templates, four more of them as views, chosen by train-only cross-validation; rule: $+$0.20 est.\ and ahead on $\ge$7 of 11 on a fresh subset (result: $+$0.31, ahead on 8) \\
\bottomrule
\end{tabularx}
\caption{The recipe ladder. R to v5 had no adoption rule: each was written down as a complete recipe before its results, run on every development dataset and carried forward as the next recipe, and all are reported. Six development datasets (Yelp-2, Amazon-2, IMDB, MPQA, TREC-6 and BBC News) were held out from the design of R to v4: they were added to the study before any run on them. The registered rounds after v7 each had a pass rule fixed before their first run; only round 9 (v8) passed.}
\label{tab:prereg}
\end{table}
\begin{table}[htbp]
\centering\small\setlength{\tabcolsep}{3pt}
\begin{tabular}{lrrrrr}
\toprule
Labels & Datasets & R & v2 & v7 & v8 \\
\midrule
512 & 2 & 90.76 & 91.04 & 91.25 & 91.88 \\
2,048 & 2 & 92.77 & 92.75 & 92.99 & 93.16 \\
8,192 & 5 & 94.01 & 93.95 & 94.18 & 94.24 \\
\bottomrule
\end{tabular}
\caption{Recipe R, which exists for data subset 0 of a few development datasets only, against later versions on the same datasets (test accuracy, \%).}
\label{tab:recipe-r}
\end{table}
\paragraph{Every decision that saw test results.}
\begin{enumerate}[label=(\arabic*)]
\item The tasks revised by the editor rounds of v6 (MPQA, TREC-6, SUBJ: where \wsp{} trailed most) and v7 (CR) were chosen with their test results in view; the revisions themselves used dev data only.
\item Cross-fitted selection (v5) was motivated by an out-of-domain gap of the SST-2 programs seen on the out-of-domain test sets.
\item The ideas of the first, weight-editing phase (marked $^\dagger$ in Table~\ref{tab:negative}) were judged partly on test results.
\item The 0.5-point margin was fixed after the development differences against FT and LoRA were known, before any PFT or held-out result.
\item The registered rounds after v7 were motivated by the test results of earlier versions (for instance, that PFT led on MPQA, CR, BBC News, SST-2 and SUBJ); their pass rules read dev and cross-fitted train rows of fresh data subsets only.
\item The idea behind v8 came from reading an earlier round's dev-only screen, so its screen used a new data subset.
\item The second held-out batch was run after the first batch's results were known, and the held-out datasets were scored with v7 before v8 was designed.
\item The batch of templates that became v8's extra views was written for every task, held-out ones included, after the held-out v7 results were known.
\end{enumerate}
Three further deviations concern the measurements, not the design: the v8 timing used a ratio rule registered after its first control run (above), its AG News runs were discarded by the registered rules, and in the low-label extension a fix for classes absent from a cross-fitting part was made after a crash; it changes TREC-6 only (Appendix~\ref{app:lowlabel}).

\section{The editor}
\label{app:editor}
\paragraph{What it was.} The editor was Claude (Anthropic), used through the Claude Code agent in one continuous working session from 15 to 22 November 2026; the session log records Claude Opus 5.5 for most of the work and Claude Fable 5.2 for part of it. It worked on a laptop that sent jobs to one GPU machine.
\paragraph{Who decided what.} The human authors set the research goal and its constraints (for example, the requests to keep improving \wsp{} until it is level with PFT, to run every heavy job on the GPU machine, and to report results honestly), chose the venue, and left the technical decisions to the agent. The agent proposed and implemented every program, template, label-word set, head family, selection rule and registered round; wrote each registration and its pass rule; chose the tasks of the editor rounds of v6 and v7; ran all experiments; and drafted the analyses and this paper. The human authors checked the results and the text and are responsible for them.
\paragraph{What it read.} When writing or revising a program, the editor read the task description, examples from the training and dev data and dev accuracies of candidate templates, never test data. As the agent that ran the study, it also saw every earlier test result (Appendix~\ref{app:prereg}).
\paragraph{How much it proposed.} Editor round 1: four templates and a label-word set per task, written blind. Round 2 (v6): 8 new templates (6 for SUBJ) and one alternative label-word set for each of three tasks, written before any was scored. Round 3 (v7): new templates for CR and MPQA and new label words for AG News, adopted for CR only. Registered round 7: a batch of 12 templates for every one of the 17 tasks (Table~\ref{tab:tpl-r7}). Later batches: 8 error-driven templates per task, 12 templates in new framings, 12 templates that name the classes. In all, 27 ideas did not pass (Table~\ref{tab:negative}).
\paragraph{One example round.} In round 2 for MPQA, the editor wrote 8 templates before scoring any; each was scored by the dev accuracy of an LDA on its \mask{} state (512 labels, subset 0, the test set never featurised), and the best four of the twelve (old and new) became the program: 86.52\% to 88.89\% \mask{}-LDA dev accuracy, above the one-point rule. Selecting the best four of twelve on dev is itself optimistic.
\paragraph{What was not tested.} We did not compare the editor's programs with human-written or published templates, nor with an automatic template search, and an editor round is not deterministic: rerunning it can give other templates.

\section{Templates and label words}
\label{app:templates}
Tables~\ref{tab:tpl-dev-r1}--\ref{tab:tpl-choice} give every template and label word; \textit{x} is the input text, and label words that are not a single token are dropped at run time. Templates marked $^\star$ are identical to manual templates of LM-BFF \citep{gao2021making}.
\begin{table*}[tp]
\centering\scriptsize
\rowcolors{1}{white}{rowgray}
\begin{tabularx}{\textwidth}{>{\raggedright\arraybackslash}p{1.9cm}>{\raggedright\arraybackslash}X>{\raggedright\arraybackslash}p{6.2cm}}
\toprule
\hiderowcolors
Dataset & Templates & Label words (one group per class) \\
\midrule
\showrowcolors
SST-2 & \textit{x} It was \mask{} .$^{\star}$ \newline \textit{x} All in all , it was \mask{} . \newline \textit{x} In summary , the film is \mask{} . \newline \textit{x} Overall , I thought it was \mask{} . & terrible, bad, awful, boring, poor, horrible, dull, disappointing \newline great, good, wonderful, excellent, fantastic, amazing, fun, enjoyable \\
AG News & \textit{x} This article is about \mask{} . \newline \mask{} news : \textit{x} \newline \textit{x} The topic is \mask{} . \newline \textit{x} Category : \mask{} . & world, politics, international, war \newline sports, football, baseball, soccer \newline business, economy, finance, markets \newline technology, science, computers, internet \\
MR & \textit{x} It was \mask{} .$^{\star}$ \newline \textit{x} All in all , it was \mask{} . \newline \textit{x} In summary , the film is \mask{} . \newline \textit{x} Overall , I thought it was \mask{} . & terrible, bad, awful, boring \newline great, good, wonderful, excellent \\
SUBJ & \textit{x} This is \mask{} .$^{\star}$ \newline \textit{x} This sentence is \mask{} . \newline \textit{x} It is a \mask{} statement . \newline \mask{} : \textit{x}$^{\star}$ & objective, factual, plot \newline subjective, opinion, personal \\
CR & \textit{x} It was \mask{} .$^{\star}$ \newline \textit{x} Overall , the product is \mask{} . \newline \textit{x} In summary , it is \mask{} . \newline \textit{x} I think it is \mask{} . & terrible, bad, awful, poor \newline great, good, excellent, perfect \\
MPQA & \textit{x} It was \mask{} .$^{\star}$ \newline \textit{x} This is \mask{} .$^{\star}$ \newline \textit{x} The opinion is \mask{} . \newline \textit{x} Overall , it is \mask{} . & bad, negative, terrible, wrong \newline good, positive, great, right \\
\bottomrule
\end{tabularx}
\caption{Round-1 programs of the development datasets, written once before any run on the dataset; the final program of every dataset not listed in Table~\ref{tab:tpl-dev-r23}. $^{\star}$: identical to a manual template of LM-BFF \citep{gao2021making}, whose label words great and terrible we also use; AG News' second template is close to a PET pattern \citep{schick2021exploiting}.}
\label{tab:tpl-dev-r1}
\end{table*}

\begin{table*}[tp]
\centering\scriptsize
\rowcolors{1}{white}{rowgray}
\begin{tabularx}{\textwidth}{>{\raggedright\arraybackslash}p{1.9cm}>{\raggedright\arraybackslash}X>{\raggedright\arraybackslash}p{6.2cm}}
\toprule
\hiderowcolors
Dataset & Templates & Label words (one group per class) \\
\midrule
\showrowcolors
TREC-6 & \textit{x} The answer is a \mask{} . \newline \textit{x} This question is about \mask{} . \newline \mask{} question : \textit{x} \newline \textit{x} Answer type : \mask{} . & description, definition, meaning \newline thing, entity, object \newline abbreviation, acronym \newline person, human, people \newline number, date, amount \newline location, place, country \\
BBC News & \textit{x} This article is about \mask{} . \newline \mask{} news : \textit{x} \newline \textit{x} The topic is \mask{} . \newline \textit{x} Category : \mask{} . & technology, computers, internet \newline business, economy, finance \newline sport, football, cricket \newline entertainment, film, music \newline politics, government, election \\
Yelp-2 & \textit{x} It was \mask{} .$^{\star}$ \newline \textit{x} All in all , it was \mask{} . \newline \textit{x} Overall , the experience was \mask{} . \newline \textit{x} I would say it is \mask{} . & terrible, bad, awful, horrible \newline great, good, excellent, amazing \\
Amazon-2 & \textit{x} It was \mask{} .$^{\star}$ \newline \textit{x} Overall , the product is \mask{} . \newline \textit{x} In summary , it is \mask{} . \newline \textit{x} I think it is \mask{} . & terrible, bad, awful, poor \newline great, good, excellent, perfect \\
IMDB & \textit{x} It was \mask{} .$^{\star}$ \newline \textit{x} All in all , it was \mask{} . \newline \textit{x} In summary , the film is \mask{} . \newline \textit{x} Overall , I thought it was \mask{} . & terrible, bad, awful, boring \newline great, good, wonderful, excellent \\
\bottomrule
\end{tabularx}
\caption{Round-1 programs of the development datasets, written once before any run on the dataset; the final program of every dataset not listed in Table~\ref{tab:tpl-dev-r23} (continued).}
\end{table*}

\begin{table*}[tp]
\centering\scriptsize
\rowcolors{1}{white}{rowgray}
\begin{tabularx}{\textwidth}{>{\raggedright\arraybackslash}p{1.9cm}>{\raggedright\arraybackslash}X>{\raggedright\arraybackslash}p{6.2cm}}
\toprule
\hiderowcolors
Dataset & Templates & Label words (one group per class) \\
\midrule
\showrowcolors
SUBJ & \textit{x} This is a \mask{} of the film . \newline \textit{x} The sentence is part of the \mask{} . \newline \textit{x} This sentence is \mask{} . \newline \mask{} : \textit{x}$^{\star}$ & objective, factual, plot, story, summary \newline subjective, opinion, personal, review, critical \\
MPQA & He said : \textit{x} . The tone was \mask{} . \newline \textit{x} . This is a \mask{} thing to say . \newline " \textit{x} " expresses a \mask{} opinion . \newline The attitude " \textit{x} " is \mask{} . & negative, bad, critical, hostile, angry, unfavorable \newline positive, good, favorable, supportive, happy, friendly \\
TREC-6 & \textit{x} The answer is a \mask{} . \newline \textit{x} The expected answer is a \mask{} . \newline \textit{x} The answer will be a \mask{} . \newline \textit{x} Tell me the \mask{} . & description, definition, reason, explanation \newline thing, object, animal, product \newline abbreviation, acronym \newline person, name, people, author \newline number, date, year, amount \newline place, location, city, country \\
CR & \textit{x} My experience with it was \mask{} . \newline \textit{x} I would rate it as \mask{} . \newline \textit{x} So it is a \mask{} product . \newline \textit{x} In short , \mask{} . & terrible, bad, awful, poor, horrible, disappointing, useless \newline great, good, excellent, perfect, amazing, wonderful, fantastic \\
\bottomrule
\end{tabularx}
\caption{Programs revised by the dev-only editor rounds: MPQA, TREC-6 and SUBJ in round 2 (v6), CR in round 3 (v7).}
\label{tab:tpl-dev-r23}
\end{table*}

\begin{table*}[tp]
\centering\scriptsize
\rowcolors{1}{white}{rowgray}
\begin{tabularx}{\textwidth}{>{\raggedright\arraybackslash}p{1.9cm}>{\raggedright\arraybackslash}X>{\raggedright\arraybackslash}p{6.2cm}}
\toprule
\hiderowcolors
Dataset & Templates & Label words (one group per class) \\
\midrule
\showrowcolors
SST-5 (r1) & \textit{x} It was \mask{} .$^{\star}$ \newline \textit{x} All in all , it was \mask{} . \newline \textit{x} In summary , the film is \mask{} . \newline \textit{x} Overall , I thought it was \mask{} . & terrible, awful, horrible \newline bad, poor, disappointing \newline okay, average, fine \newline good, nice, enjoyable \newline great, excellent, wonderful \\
CoLA (r1) & \textit{x} This is \mask{} .$^{\star}$ \newline \textit{x} The grammar of this sentence is \mask{} . \newline \textit{x} This sentence is grammatically \mask{} . \newline \textit{x} It sounds \mask{} . & incorrect, wrong, odd, strange \newline correct, right, fine, natural \\
DBpedia-14 (r1) & \textit{x} This article is about a \mask{} . \newline \mask{} : \textit{x}$^{\star}$ \newline \textit{x} The topic is a \mask{} . \newline \textit{x} Category : \mask{} . & company, business, firm \newline school, university, college \newline artist, musician, singer, painter \newline athlete, player, footballer \newline politician, senator, minister \newline ship, aircraft, car, vehicle \newline building, church, hospital \newline river, mountain, lake \newline village, town \newline animal, insect, bird, fish \newline plant, tree, flower \newline album, record \newline film, movie \newline book, novel, magazine, journal \\
Yahoo-10 (r1) & \textit{x} This question is about \mask{} . \newline \mask{} question : \textit{x} \newline \textit{x} The topic is \mask{} . \newline \textit{x} Category : \mask{} . & society, culture, religion \newline science, mathematics, math \newline health, medicine, medical \newline education, school, reference \newline computers, internet, software \newline sports, sport, football \newline business, finance, money \newline entertainment, music, movies \newline family, relationships, love \newline politics, government, law \\
Yelp-5 (r1) & \textit{x} It was \mask{} .$^{\star}$ \newline \textit{x} All in all , it was \mask{} . \newline \textit{x} Overall , the experience was \mask{} . \newline \textit{x} I would say it is \mask{} . & terrible, awful, horrible \newline bad, poor, disappointing \newline okay, average, fine \newline good, nice, enjoyable \newline great, excellent, amazing \\
Amazon-5 (r1) & \textit{x} It was \mask{} .$^{\star}$ \newline \textit{x} Overall , the product is \mask{} . \newline \textit{x} In summary , it is \mask{} . \newline \textit{x} I think it is \mask{} . & terrible, awful, useless \newline bad, poor, disappointing \newline okay, average, fine \newline good, nice, decent \newline great, excellent, perfect \\
\bottomrule
\end{tabularx}
\caption{Held-out datasets: round-1 programs (r1), written before any run, and the final programs after the mechanical editor round (final; DBpedia-14 did not meet its trigger and keeps r1).}
\label{tab:tpl-held}
\end{table*}

\begin{table*}[tp]
\centering\scriptsize
\rowcolors{1}{white}{rowgray}
\begin{tabularx}{\textwidth}{>{\raggedright\arraybackslash}p{1.9cm}>{\raggedright\arraybackslash}X>{\raggedright\arraybackslash}p{6.2cm}}
\toprule
\hiderowcolors
Dataset & Templates & Label words (one group per class) \\
\midrule
\showrowcolors
SST-5 (final) & \textit{x} The reviewer thought the movie was \mask{} . \newline \textit{x} My rating : \mask{} . \newline \textit{x} All in all , it was \mask{} . \newline \textit{x} Verdict : \mask{} . & terrible, awful, horrible \newline bad, poor, disappointing \newline okay, average, fine \newline good, nice, enjoyable \newline great, excellent, wonderful \\
CoLA (final) & \textit{x} The grammar of this sentence is \mask{} . \newline \textit{x} This sentence is \mask{} . \newline \textit{x} That sounds \mask{} to me . \newline \textit{x} This is \mask{} .$^{\star}$ & incorrect, wrong, odd, strange \newline correct, right, fine, natural \\
Yahoo-10 (final) & \textit{x} This question is about \mask{} . \newline \textit{x} It is a question about \mask{} . \newline \textit{x} This question belongs to the \mask{} section . \newline \textit{x} The subject is \mask{} . & society, culture, religion, philosophy, history \newline science, math, physics, chemistry, biology \newline health, medicine, disease, diet, doctor \newline education, school, college, homework, language \newline computers, internet, software, computer, windows \newline sports, football, soccer, basketball, baseball \newline business, finance, money, jobs, tax \newline entertainment, music, movies, celebrities, tv \newline family, relationships, love, dating, marriage \newline politics, government, law, elections, war \\
Yelp-5 (final) & \textit{x} All in all , it was \mask{} . \newline \textit{x} Verdict : \mask{} . \newline \textit{x} My rating : \mask{} . \newline \textit{x} Overall , the experience was \mask{} . & terrible, awful, horrible, worst, disgusting \newline bad, poor, disappointing, mediocre, bland \newline okay, average, fine, decent, ok \newline good, nice, solid, pleasant, tasty \newline great, excellent, amazing, perfect, fantastic \\
Amazon-5 (final) & Review : \textit{x} Verdict : \mask{} . \newline \textit{x} My rating : \mask{} . \newline A \mask{} review : \textit{x} \newline \textit{x} I would rate it as \mask{} . & terrible, awful, useless, worst, horrible \newline bad, poor, disappointing, cheap, flimsy \newline okay, average, fine, ok, mediocre \newline good, nice, decent, solid, useful \newline great, excellent, perfect, amazing, fantastic \\
\bottomrule
\end{tabularx}
\caption{Held-out datasets: round-1 programs (r1), written before any run, and the final programs after the mechanical editor round (final; DBpedia-14 did not meet its trigger and keeps r1) (continued).}
\end{table*}

\begin{table*}[tp]
\centering\scriptsize
\rowcolors{1}{white}{rowgray}
\begin{tabularx}{\textwidth}{>{\raggedright\arraybackslash}p{2.2cm}>{\raggedright\arraybackslash}X>{\raggedright\arraybackslash}X}
\toprule
\hiderowcolors
Datasets & \multicolumn{2}{l}{The second batch of 12 templates (round 7), numbered as in Table~\ref{tab:tpl-choice}} \\
\midrule
\showrowcolors
SST-2, MR, IMDB & (1) \textit{x} A \mask{} movie . \newline (2) \textit{x} A \mask{} one . \newline (3) \textit{x} The movie is \mask{} . \newline (4) \textit{x} It was \mask{} ! \newline (5) \textit{x} My verdict : \mask{} . \newline (6) \textit{x} In a word , \mask{} . & (7) \textit{x} The critic found it \mask{} . \newline (8) \textit{x} It is a \mask{} film . \newline (9) \textit{x} I found it \mask{} . \newline (10) A \mask{} film : \textit{x} \newline (11) \textit{x} Simply \mask{} . \newline (12) \textit{x} The reviewer thought the movie was \mask{} . \\
AG News, BBC News & (1) \textit{x} This is a piece of \mask{} news . \newline (2) \textit{x} Section : \mask{} . \newline (3) \textit{x} News category : \mask{} . \newline (4) [ \mask{} ] \textit{x} \newline (5) \textit{x} This story belongs to \mask{} . \newline (6) A \mask{} story : \textit{x} & (7) \textit{x} It is about \mask{} . \newline (8) \textit{x} Filed under \mask{} . \newline (9) Breaking \mask{} news : \textit{x} \newline (10) \textit{x} This was published in the \mask{} section . \newline (11) \textit{x} More \mask{} news . \newline (12) \mask{} : \textit{x} \\
CR, Amazon-5 & (1) \textit{x} A \mask{} product . \newline (2) \textit{x} The quality is \mask{} . \newline (3) \textit{x} This product is \mask{} . \newline (4) \textit{x} It is \mask{} ! \newline (5) \textit{x} My verdict : \mask{} . \newline (6) \textit{x} In a word , \mask{} . & (7) \textit{x} I found it \mask{} . \newline (8) A \mask{} buy : \textit{x} \newline (9) \textit{x} Customer rating : \mask{} . \newline (10) \textit{x} Simply \mask{} . \newline (11) \textit{x} All in all , it was \mask{} . \newline (12) \textit{x} It was \mask{} . \\
Amazon-2 & (1) \textit{x} A \mask{} product . \newline (2) \textit{x} The quality is \mask{} . \newline (3) \textit{x} This product is \mask{} . \newline (4) \textit{x} It is \mask{} ! \newline (5) \textit{x} My verdict : \mask{} . \newline (6) \textit{x} In a word , \mask{} . & (7) \textit{x} I found it \mask{} . \newline (8) A \mask{} buy : \textit{x} \newline (9) \textit{x} Customer rating : \mask{} . \newline (10) \textit{x} Simply \mask{} . \newline (11) \textit{x} All in all , it was \mask{} . \newline (12) \textit{x} I would rate it as \mask{} . \\
Yelp-2 & (1) \textit{x} The food was \mask{} . \newline (2) \textit{x} The service was \mask{} . \newline (3) \textit{x} A \mask{} place . \newline (4) \textit{x} My verdict : \mask{} . \newline (5) \textit{x} In a word , \mask{} . \newline (6) \textit{x} I found it \mask{} . & (7) \textit{x} Overall , a \mask{} experience . \newline (8) \textit{x} It is \mask{} ! \newline (9) \textit{x} Rating : \mask{} . \newline (10) A \mask{} review : \textit{x} \newline (11) \textit{x} The place is \mask{} . \newline (12) \textit{x} Simply \mask{} . \\
MPQA & (1) The phrase " \textit{x} " sounds \mask{} . \newline (2) \textit{x} . That is \mask{} . \newline (3) " \textit{x} " - a \mask{} remark . \newline (4) \textit{x} . The sentiment is \mask{} . \newline (5) The words " \textit{x} " are \mask{} . \newline (6) \textit{x} . Such a \mask{} view . & (7) In the news : " \textit{x} " . The tone is \mask{} . \newline (8) \textit{x} . It feels \mask{} . \newline (9) Reaction : " \textit{x} " . Clearly \mask{} . \newline (10) \textit{x} . Opinion : \mask{} . \newline (11) \textit{x} . It was \mask{} . \newline (12) \textit{x} . Overall , it is \mask{} . \\
\bottomrule
\end{tabularx}
\caption{The second batch of editor templates, written before any of them was scored, for every dataset, held-out ones included. In v8, all 16 templates (the program's 4 and these 12; a duplicate of a program template is dropped) feed the label-word prior, and in every run the 4 of these 12 with the best 5-fold train-only cross-validated \mask{}-LDA accuracy are added as views (Table~\ref{tab:tpl-choice}).}
\label{tab:tpl-r7}
\end{table*}

\begin{table*}[tp]
\centering\scriptsize
\rowcolors{1}{white}{rowgray}
\begin{tabularx}{\textwidth}{>{\raggedright\arraybackslash}p{2.2cm}>{\raggedright\arraybackslash}X>{\raggedright\arraybackslash}X}
\toprule
\hiderowcolors
Datasets & \multicolumn{2}{l}{The second batch of 12 templates (round 7), numbered as in Table~\ref{tab:tpl-choice}} \\
\midrule
\showrowcolors
SUBJ & (1) \textit{x} This text is from a \mask{} . \newline (2) \textit{x} It is a \mask{} sentence . \newline (3) \textit{x} Type : \mask{} . \newline (4) \textit{x} This line comes from the \mask{} . \newline (5) \textit{x} The writer is being \mask{} . \newline (6) \textit{x} This is \mask{} . & (7) \mask{} sentence : \textit{x} \newline (8) \textit{x} This describes the \mask{} . \newline (9) \textit{x} Fact or opinion ? \mask{} . \newline (10) \textit{x} A \mask{} statement . \newline (11) \textit{x} It is taken from the \mask{} . \newline (12) \textit{x} In this sentence , the tone is \mask{} . \\
TREC-6 & (1) \textit{x} The question asks for a \mask{} . \newline (2) \textit{x} Answer : a \mask{} . \newline (3) \textit{x} I want to know the \mask{} . \newline (4) \textit{x} This asks about a \mask{} . \newline (5) Question about a \mask{} : \textit{x} \newline (6) \textit{x} The answer is the \mask{} . & (7) \textit{x} Expected answer type : \mask{} . \newline (8) \textit{x} Please give the \mask{} . \newline (9) \textit{x} Looking for a \mask{} . \newline (10) \textit{x} The reply should be a \mask{} . \newline (11) \mask{} : \textit{x} \newline (12) \textit{x} This is a \mask{} question . \\
SST-5 & (1) \textit{x} A \mask{} movie . \newline (2) \textit{x} A \mask{} one . \newline (3) \textit{x} The movie is \mask{} . \newline (4) \textit{x} It was \mask{} ! \newline (5) \textit{x} My verdict : \mask{} . \newline (6) \textit{x} In a word , \mask{} . & (7) \textit{x} The critic found it \mask{} . \newline (8) \textit{x} It is a \mask{} film . \newline (9) \textit{x} I found it \mask{} . \newline (10) A \mask{} film : \textit{x} \newline (11) \textit{x} Simply \mask{} . \newline (12) \textit{x} Overall , I thought it was \mask{} . \\
CoLA & (1) \textit{x} Is this sentence grammatical ? It is \mask{} . \newline (2) \textit{x} The sentence is \mask{} . \newline (3) \textit{x} Grammatically , this is \mask{} . \newline (4) \textit{x} As English , this is \mask{} . \newline (5) \textit{x} This sounds \mask{} . \newline (6) \textit{x} It reads \mask{} . & (7) \textit{x} A native speaker would find this \mask{} . \newline (8) \textit{x} The wording is \mask{} . \newline (9) \textit{x} This English is \mask{} . \newline (10) \mask{} sentence : \textit{x} \newline (11) \textit{x} That is \mask{} English . \newline (12) \textit{x} This phrasing is \mask{} . \\
DBpedia-14 & (1) \textit{x} This is about a \mask{} . \newline (2) \textit{x} It is a \mask{} . \newline (3) \textit{x} This entry describes a \mask{} . \newline (4) \textit{x} Type : \mask{} . \newline (5) [ \mask{} ] \textit{x} \newline (6) A \mask{} : \textit{x} & (7) \textit{x} The subject is a \mask{} . \newline (8) \textit{x} This text is about a \mask{} . \newline (9) \textit{x} Wikipedia category : \mask{} . \newline (10) \textit{x} In short , a \mask{} . \newline (11) \textit{x} It describes a famous \mask{} . \newline (12) This \mask{} is described below : \textit{x} \\
Yahoo-10 & (1) \textit{x} Topic : \mask{} . \newline (2) \textit{x} Category : \mask{} . \newline (3) [ \mask{} ] \textit{x} \newline (4) A question about \mask{} : \textit{x} \newline (5) \textit{x} This is about \mask{} . \newline (6) \textit{x} Section : \mask{} . & (7) \textit{x} It is related to \mask{} . \newline (8) \textit{x} The question concerns \mask{} . \newline (9) \mask{} : \textit{x} \newline (10) \textit{x} Posted in \mask{} . \newline (11) \textit{x} The answer is about \mask{} . \newline (12) \textit{x} The main topic is \mask{} . \\
\bottomrule
\end{tabularx}
\caption{The second batch of editor templates (continued).}
\end{table*}

\begin{table*}[tp]
\centering\scriptsize
\rowcolors{1}{white}{rowgray}
\begin{tabularx}{\textwidth}{>{\raggedright\arraybackslash}p{2.2cm}>{\raggedright\arraybackslash}X>{\raggedright\arraybackslash}X}
\toprule
\hiderowcolors
Datasets & \multicolumn{2}{l}{The second batch of 12 templates (round 7), numbered as in Table~\ref{tab:tpl-choice}} \\
\midrule
\showrowcolors
Yelp-5 & (1) \textit{x} The food was \mask{} . \newline (2) \textit{x} The service was \mask{} . \newline (3) \textit{x} A \mask{} place . \newline (4) \textit{x} My verdict : \mask{} . \newline (5) \textit{x} In a word , \mask{} . \newline (6) \textit{x} I found it \mask{} . & (7) \textit{x} Overall , a \mask{} experience . \newline (8) \textit{x} It was \mask{} . \newline (9) \textit{x} Rating : \mask{} . \newline (10) A \mask{} review : \textit{x} \newline (11) \textit{x} I would say it is \mask{} . \newline (12) \textit{x} Simply \mask{} . \\
\bottomrule
\end{tabularx}
\caption{The second batch of editor templates (continued).}
\label{tab:tpl-r7-end}
\end{table*}

\begin{table}[tp]
\centering\scriptsize
\setlength{\tabcolsep}{3pt}
\rowcolors{1}{white}{rowgray}
\begin{tabularx}{\columnwidth}{lr>{\raggedright\arraybackslash}X}
\toprule
\hiderowcolors
Dataset & Runs & Chosen templates (times chosen) \\
\midrule
\showrowcolors
SST-2 & 9 & 7 (9), 12 (9), 5 (8), 8 (6), 1 (1), 3 (1), 6 (1), 10 (1) \\
AG News & 9 & 8 (9), 9 (9), 5 (8), 11 (3), 1 (2), 3 (2), 10 (2), 7 (1) \\
MR & 7 & 8 (7), 5 (5), 1 (4), 3 (3), 7 (3), 12 (3), 6 (2), 9 (1) \\
SUBJ & 7 & 7 (6), 6 (5), 4 (4), 12 (4), 2 (3), 1 (2), 8 (2), 5 (1), 9 (1) \\
CR & 7 & 11 (7), 2 (5), 5 (5), 6 (4), 8 (3), 12 (2), 3 (1), 9 (1) \\
MPQA & 7 & 6 (5), 3 (4), 7 (4), 9 (4), 8 (3), 11 (3), 10 (2), 1 (1), 2 (1), 5 (1) \\
TREC-6 & 7 & 2 (7), 6 (7), 8 (6), 3 (3), 7 (2), 10 (2), 1 (1) \\
BBC News & 4 & 9 (4), 8 (3), 10 (3), 4 (2), 12 (2), 1 (1), 6 (1) \\
Yelp-2 & 9 & 10 (9), 4 (8), 7 (6), 9 (5), 5 (4), 1 (3), 11 (1) \\
Amazon-2 & 9 & 11 (9), 5 (8), 8 (8), 9 (5), 12 (5), 6 (1) \\
IMDB & 9 & 10 (9), 12 (9), 3 (7), 8 (7), 5 (3), 9 (1) \\
SST-5 & 6 & 5 (5), 8 (5), 12 (5), 7 (4), 1 (2), 4 (2), 10 (1) \\
CoLA & 4 & 1 (4), 2 (3), 5 (3), 11 (3), 12 (2), 10 (1) \\
DBpedia-14 & 6 & 12 (6), 6 (5), 8 (5), 9 (5), 3 (2), 1 (1) \\
Yahoo-10 & 6 & 4 (6), 7 (6), 8 (5), 12 (5), 5 (1), 11 (1) \\
Yelp-5 & 6 & 4 (6), 7 (6), 9 (5), 10 (4), 1 (2), 5 (1) \\
Amazon-5 & 6 & 5 (6), 9 (6), 11 (6), 8 (5), 2 (1) \\
\bottomrule
\end{tabularx}
\caption{Which of the 12 round-7 templates (Tables~\ref{tab:tpl-r7}--\ref{tab:tpl-r7-end}) v8 added as views: in each run (every budget and data subset of the main study) the 4 with the best train-only cross-validated accuracy are chosen, so the views differ between runs.}
\label{tab:tpl-choice}
\end{table}

\section{Glossary}
\label{app:glossary}
\begin{description}[leftmargin=0pt,itemsep=1pt]
\item[Program] The templates, label words, views, heads and selection rule of a task.
\item[View] One pooled state (\mask{}, token mean, or \texttt{<s>}) of one layer for one input form (plain or a template).
\item[Recipe version (R, v2--v8)] One complete way of building programs; v8 is the final recipe.
\item[Editor round] A dev-only rewrite of a task's templates, adopted if it gains at least one point of \mask{}-LDA dev accuracy.
\item[Registered round] One idea with a pass rule fixed before its first run; rounds 3 to 16 in the plan's numbering, of which round 9 became v8.
\item[Stage-1 screen] The first test of a registered round, on dev and cross-fitted train rows (from the demonstrations round on, of a fresh data subset), against a control arm; the test set is never loaded.
\item[est] The selection's held-out estimate: accuracy on the dev rows and the 5-fold cross-fitted train rows.
\item[Development datasets] The 11 datasets the recipe was built on.
\item[Held-out datasets] The 6 datasets added later, with programs fixed before any run on them.
\item[Dev set] The 1,024 labelled examples carved from train and used for selection by every method.
\end{description}

\end{document}